\documentclass[11pt]{article}
\usepackage[utf8]{inputenc}
\usepackage{graphics}
\usepackage[a4paper, total={7.1in, 10.1in}]{geometry}
\usepackage{geometry} 
\renewcommand{\baselinestretch}{1.5}
\usepackage[english]{babel}
\usepackage{pstricks}
\usepackage{pst-node}
\usepackage{amsmath}
\usepackage{authblk}
\usepackage{url}
\usepackage{tikz}
\usetikzlibrary{backgrounds,patterns,shapes}
\usepackage{booktabs}
\usepackage[natbibapa]{apacite}
\usepackage{multirow}
\usepackage{subcaption}
\usepackage{caption} 
\usepackage{float}
\usepackage{xcolor}
\usepackage{enumitem}

\usepackage{eurosym}

\usepackage{color}

\usepackage[markup=underlined]{changes}
\definechangesauthor[color=BrickRed]{ME}

\providecommand{\keywords}[1]
{
  \textbf{\textit{Keywords---}} #1
}

\title{\vspace{2cm} \Large \textbf{ Modeling and solving the multimodal car- and ride-sharing problem }}
\bigskip
\author[a,b]{\bf Miriam Enzi \footnote{Corresponding author.}}
\author[b]{\bf Sophie N. Parragh}
\author[c]{\bf David Pisinger}
\author[a]{\bf Matthias Prandtstetter}
\bigskip
\affil[a]{Center for Mobility Systems, AIT Austrian Institute of Technology \protect\\
Giefinggasse 4, 1210 Vienna, Austria \protect\\
\texttt{matthias.prandtstetter@ait.ac.at}
\medskip }
\affil[b]{Institute of Production and Logistics Management, Johannes Kepler University Linz \protect\\
Altenberger Straße 69, 4040 Linz, Austria \protect\\
\texttt{\{miriam.enzi,sophie.parragh\}@jku.at}
\medskip }
\affil[c]{Department of Management Engineering, Technical University of Denmark \protect\\
Akademivej Building 358, 2800 Kgs. Lyngby, Denmark \protect\\
\texttt{pisinger@man.dtu.dk}
\medskip }

\newcommand{\myemph}[1]      {{\it #1}}

\newcommand{\dualalpha}      {y}
\newcommand{\dualbeta}       {u}
\newcommand{\dualgamma}      {\overline{u}}
\newcommand{\myG}            {F}
\newcommand{\myH}            {G}
\newcommand{\myQ}            {H}

\newcommand{\heuredges}      {\texttt{heurarcs}}
\newcommand{\heurprun}       {\texttt{heurprun}}
\newcommand{\statespace}     {\texttt{statespace}}

\newcommand{\best}           {\texttt{best}}
\newcommand{\first}          {\texttt{first}}
\newcommand{\firstdep}       {\texttt{firstdep}}
\newcommand{\multiple}       {\texttt{multiple}}

\begin{document}

\begin{titlepage}
\maketitle
\thispagestyle{empty}
\end{titlepage}

\section*{Abstract}
We introduce the multimodal car- and ride-sharing problem (MMCRP), in which a pool of cars is used to cover a set of ride requests while uncovered requests are assigned to other modes of transport (MOT). A car’s route consists of one or more trips. Each trip must have a specific but non-predetermined driver, start in a depot and finish in a (possibly different) depot. Ride-sharing between users is allowed, even when two rides do not have the same origin and/or destination. A user has always the option of using other modes of transport according to an individual list of preferences.

The problem can be formulated as a vehicle scheduling problem. In order to solve the problem, an auxiliary graph is constructed in which each trip starting and ending in a depot, and covering possible ride-shares, is modeled as an arc in a time-space graph. We propose a two-layer decomposition algorithm based on column generation, where the master problem ensures that each request can only be covered at most once, and the pricing problem generates new promising routes by solving a kind of shortest-path problem in a time-space network. 
Computational experiments based on realistic instances are reported. The benchmark instances are based on demographic, spatial, and economic data of Vienna, Austria.  
We solve large instances with the column generation based approach to near optimality in reasonable time, and we further investigate various exact and heuristic pricing schemes.

\bigskip
\keywords{transportation, car-sharing, ride-sharing, vehicle scheduling problem, column generation}

\section{Introduction} 

Studying the development of mobility during the last decades, one can easily observe that we are facing a strong wind of change. While some years ago the main developments were related to technological improvements, the introduction of e-mobility was a first step towards an ongoing kind of revolution. From there, a new understanding of mobility developed and we are now facing a future where ``owning cars'' is replaced by ``being mobile''. Users preferably only specify cornerstones of their travel --- such as origins and destinations, latest arrival times or preferable modes of transport (MOT) --- and rely on an information system to provide an (optimal) assignment of modes of transport to their demands. This attitude is supported by mobility concepts like vehicle sharing (car and bike), easy access to mobility via mobility cards, and Mobility as a Service \citep{Mulley2018}. We observe these developments not only in the private sector but also in the area of corporate mobility. Increasingly, companies are trying to change their view on their corporate mobility by switching from individually assigned cars towards Mobility as a Service for their employees. Companies strive to have an overall green and sustainable profile and employees are aware of the importance to contribute to a greener world, even if their travel time of a trip might increase. Instead of supporting further developments in corporate mobility privileging a few selected users, we are aiming at providing sustainable corporate mobility concepts that ensure at least the same level of mobility, while increasing positive impacts (e.g., cost reduction, ecological sustainability, and employee satisfaction). 

This work is part of an applied research project \textit{SEAMLESS} (http://www.seamless-project.at), in which the project partners are implementing the discussed ideas including the supporting algorithms in their companies. The major goal of the project is the development of novel corporate mobility concepts aiming at providing mobility to the company (and its employees) instead of only providing cars. This includes the introduction of car pools that can be used by the employees on a smart assignment strategy. Additional modes of transport are incorporated like bikes, (public) bike- and car-sharing, public transport and users can co-ride with each other. Ride-sharing saves resources, such as cars and energy, it is considered to have a good environmental footprint and can solve congestion problems. \citet{Masoud2017} report that for private cars in the US with four seats, only around 1.7 seats are actually used on average. This number decreases to only 1.2 for work-based trips, which shows the underutilization of cars, especially company cars. The increasing number of empty seats in cars and an increasing number of users asking for rides, imply motivation to elaborate a sophisticated ride-sharing system. Furthermore, not only the sharing economy is increasing but also a combined and integrated use of various modes of transport. To avoid pollution and congestion problems, various cities give incentives to use (a combination of) ''greener'' modes of transport \citep{step2025,madrid}. As transportation is one of the biggest producers of emissions \citep{EUComm2016}, it is vital to consider sustainability aspects, shift to more sustainable modes and enhance the environmental footprint. 

The package of mobility offers can be seen as an extended car pool. It is crucial to assign the right vehicle to the right mobility need --- e.g., if someone is aiming to travel a short distance in the city, public transport is better suited than a conventionally driven minivan. In return, the minivan is the right choice if an employee has to transport some special equipment to a meeting at a location about 300 kilometers away. This implies that it is necessary to estimate the mobility demand, allowing the user to specify preferred modes of transport, and to determine the number (and types) of cars to be owned in the car pool as well as the mobility offers provided to the employees like mobility cards or access to public vehicle sharing systems. Although sharing reduces costs and environmental impact, the complexity in fleet management increases. This directly implies that computational support is necessary to be able to efficiently handle the fleet.

We study the multimodal car- and ride-sharing problem in a company having one or more offices from where the employees have to visit various customers during office hours (e.g., for business meetings). We consider a fixed and unique employee-to-meeting assignment and a fixed latest arrival time. This results in a fixed sequence of tasks (also referred to as trip) for every employee with several stops, starting and ending at predefined (but possibly different) depots. 
A pool of vehicles is provided to the employees (also referred to as users) who can jointly use them (car-sharing). Furthermore, up to two users may co-ride on specific legs or routes with each other (ride-sharing).

We model the trips as arcs in a directed acyclic graph. Vehicle routes consist of one or more trips, whereas the driver of these trips may change at the depot.
Similar to the vehicle scheduling problem, the available vehicles cover the scheduled trips resulting in vehicle routes. As the pool of cars is restricted, only a subset of the trips will be covered by the shared vehicles. In order to cover all mobility requests in the best possible way, further MOTs such as bikes or public transport are used. If a trip is not covered by car, the cheapest other MOT will be used. 

This paper and the project objectives focus on adapting future mobility considerations to a corporate setting. However, the results can easily be adapted for different closed groups with a predefined set of users, such as home communities, suburban areas, or simply a network of users with predefined locations where the cars must be picked-up at and returned to. Furthermore, the model can easily be adapted to bikes, segways, cargobikes, (electric) scooters, and other sharing offers.

As the problem is modeled as an extended vehicle scheduling problem (VSP) with multiple depots, we contribute to the body of this specific problem too. The VSP assigns a set of vehicles to a set of scheduled trips, such that costs are minimized and each trip is covered by exactly one vehicle \citep{Baita2000}. There are three main differences between the vehicle scheduling problem and the MMCRP. First, in our case not all trips need to be covered. Second, the multi-depot VSP assumes that all vehicles return to their original depot. We allow for different, but predetermined, start and end points. Third, we have to avoid that users co-ride in parallel on different trips. Hence we add a tailored ride-sharing constraint to the model. The MMCRP can be transformed into a kind of vehicle scheduling problem by having infinite cost for all other modes of transport, forcing the solution to cover all trips and allowing for different start and end depots. The MMCRP can also be formulated as a VSP with profits, which only  - related to the idea of the vehicle routing problem (VRP) with profits - covers trips that are profitable.

The contributions of this paper are as follows:
\begin{itemize}[noitemsep]
\item We introduce the novel MMCRP derived from a real-world application, and formulate it as an extended vehicle scheduling problem. To the best of our knowledge, it is one of the first models including both car- and ride-sharing. We show that this real-world application can be efficiently solved with well-known methods.
\item We present a two-layer decomposition of the problem. In the first layer, trips starting and ending at a depot are enumerated. The trips also take care of enumerating all possible ride-sharing possibilities. With this we are able to hide complicated constraints considering ride-sharing and make it usable for practical purpose. The framework is flexible and allows the introduction of additional constraints like detour constraints, co-riding preferences, and driving time constraints. In the second layer, the trips are combined into vehicle-routes. The second-layer decomposition is solved through a column generation based algorithm. We present an efficient algorithm for solving the pricing problem using a label setting algorithm on a directed acyclic graph (DAG). Several different pricing strategies are presented and compared. These include adding more columns in each iteration, and various heuristics in combination with an exact approach for solving the pricing problem.
\item Computational results confirm that large instances can be solved to near-optimality in reasonable time using a column generation based approach, making it possible to use the algorithm for daily planning of multimodal car- and ride-sharing systems. We also show that the gap between the LP-bound found through column generation and the integer solution obtained on the same columns is very small.
\end{itemize}

The paper is organized as follows:
First, in Section~\ref{sec:lit}, we discuss related work focusing on car- and ride-sharing as well as the VSP.
Then, we provide a detailed problem description of the MMCRP in Section~\ref{sec:probform}.
The solution approach and the auxiliary 
graph that is used in our algorithm are described in Section~\ref{sec:solutionapproach}.
Based on the auxiliary graph we present a direct formulation of the MMCRP
in Section~\ref{sec:compact}. 
In Section~\ref{sec:path}, we present a path formulation of the problem and show how its linear relaxation can be solved through delayed column generation in Section~\ref{sec:delayed}. In Section~\ref{sec:pricing} we explain how the
pricing problem can be solved through dynamic programming and propose 
various heuristics for improving the computational effort.
Section~\ref{sec:comp} presents the computational experiments.
The paper is concluded in Section~\ref{sec:concl} by summing up the achieved 
results, and proposing ideas for future research.

\section{Related work}
\label{sec:lit}

Recently, car- and ride-sharing have received considerable attention, and several variants of the problem have been studied. \citet{MOURAD2019323}
provide a thorough overview on models and algorithms for optimizing shared mobility. In the following section, we review closely related problems, including car-sharing, ride-sharing, and the vehicle scheduling problem focusing on column generation based approaches. 

\subsection{Car-sharing} 
Car-sharing systems involve a pool of cars that are shared among a set of users, who are usually known in advance (in public car-sharing systems these would be subscribers, in our case employees). The MMCRP without ride-sharing reduces to a car-sharing problem. \citet{Jor2013} and \citet{Brandstatter2016} review car-sharing optimization problems in detail. Most optimization studies consider publicly available systems and focus on rather strategic problems. In our setting we consider a car-sharing system available to a closed community only and focus on planning the daily operations.  Many studies focus on public car-sharing systems and tackle problems such as charging station placement \citep{Boy2015,Bra2017} or relocation of cars between stations \citep{Kek2006,Boy2015}.
Latest works increasingly also tackle the operational characteristics of car-sharing systems such as the effect of temporal and spatial flexibility on the performance of one-way electric car-sharing systems \citep{Boyaci2019}, integrating pickups and deliveries on shared vehicle routes \citep{Bergmann2020}, or dynamic relocation policies \citep{Repux2019}. 

\subsection{Ride-sharing} 
Ride-sharing describes co-riding of one or more users between an origin and a destination or sub-paths of it. This is also the main idea of the MMCRP, where we do not only exploit the various MOTs in the best possible way, but try to merge rides by allowing ride-sharing if it is beneficial. In the following we review some related studies addressing ride-sharing.

Dial-a-ride problems (DARP) or the closely related pick-up-and delivery problem (PDP) are often used to formulate ride-sharing activities \citep{Hos2014,Li2014}. Related to the MMCRP is also PDP with transfer \citep{Masson2014,Qu2012,Cortes2010}. An exhaustive review of these problems can be found in \citet{Ho2018}.

\citet{Masoud2017} propose a decomposition algorithm to solve a many-to-many ride-matching problem to optimality in a time-expanded network. Participants only provide the origin, destination and latest/earliest times, which is similar to our problem statement. In contrast to our problem, they strictly split riders and drivers.
\citet{Huang2016} formulate a two-stage problem minimizing total cost for long-term car-pooling. Drivers are selected, passengers assigned, and for each driver a traveling salesman problem (TSP) is solved considering constraints regarding fairness and preferences. 
\citet{BitMon2013} compute a driver's and passenger's individual paths including the mutual sub-path between two (to be determined and synchronized) points. Mutual trips are followed by their individual paths towards the driver's and passenger's destination. As in our work, they also include public transport and walking before/after ride-sharing, however the focus of the work is to determine the optimal pick-up and drop-off locations for requests.

A number of works study commuter trips \citep{Baldacci2004, Knapen2014, Regue2016}, whereas we focus on trips during working hours from/to meetings with customers. 
\citet{Chen2016} aim at minimizing the cost of commuters and business traffic of a company, which consists of the cost incurred from vehicle miles and the costs of penalizing the efficiency losses (arriving too late at meetings, waiting time for transfers, inconvenience and risk with transfers). A constructive heuristic based on savings in miles driven and cars used is introduced. The problem definition is closely related to ours, as not only commuting trips are considered but also business traffic, i.e., travels between meetings. Moreover, they also employ savings as a objective but only use a heuristic approach to solve the problem.

Contrary to other papers, we model our compact problem as a kind of vehicle scheduling problem defined on an acyclic time-space graph, where we do not model pick-ups and deliveries explicitly, but enumerate all possible ride-shares in an auxiliary graph which is used as input to the second stage model.

\subsection{Vehicle scheduling problem}
The VSP received increasing attention in the early 80s \citep{Bodin1981,Bodin1983}, and is mainly applied to time-tabled trips of public transport or crew scheduling. In the following we give a short overview on recent works on the multi-depot variant of the problem (MDVSP) using column generation based approaches to solve it. Further works elaborate the idea of the MDVSP by introducing alternative-fuel vehicles \citep{Adler2016} or considering a heterogeneous fleet of vehicles \citep{Guedes2015}. An overview of basic vehicle scheduling models is given in \citet{Bunte2009}. The MMCRP can be seen as a MDVSP with profits. The literature on the VRP with profits or closely related Orienteering Problem is vast \citep{Speranza2014,Gunawan2016}. We could not find any publications on the VSP with profits.

The MDVSP was proven to be NP-hard by \citet{Bertossi1987}.
Column generation was first applied to the MDVSP by \citet{Ribeiro1994} and extended by \citet{Hajdar2006} and \citet{Groiez2013}. Note that these algorithms focus on proven integer optimality of the entire problem, which is not the case in our work.
\citet{Pepin2008} compare five heuristic for the MDVSP and conclude that the column generation heuristic performs best assuming enough computational time is available and stability is required. 
\citet{Guedes2016} propose a simple and efficient heuristic approach for the MDVSP. The heuristic first applies state space reductions to reduce complexity and thereafter a truncated column generation approach. 
\citet{Kulkarni2018} present a new inventory formulation for the MDVSP and a column generation based heuristic proposing a novel decomposition.
The multi-depot vehicle scheduling with controlled trip shifting \citep{Desfontaines2018} is closely related to the MMCRP. The generalization of the MDVSP allows for slight modification of one trip scheduled time. Trips are multiplied, representing each trip for different starting times. The aim is to find a set of bus schedules that covers every trip exactly once by satisfying vehicle availability and minimize costs. The work introduces a two-phase matheuristic where column generation solving the linear relaxation is embedded in a diving heuristic to derive an integer solution. The sequence of trips is fixed in the first phase by the column generation approach and thereafter the copies of a trip {are} chosen using a mixed integer program. 

Note that all of the above works use either variable fixing or rounding strategies in their approaches. We solve the linear relaxation to optimality by column generation and find the integer solution by solving the original model using the obtained columns.
Furthermore, we do not tackle the standard VSP but {a kind of} MDVSP with profits in which only beneficial arcs are covered by a vehicle.
The study by \citet{Oukil2007}
is structurally similar to ours, but having some important differences. \citet{Oukil2007} focus on the comparison of the standard and the stabilized column generation approach, and discuss the impact of different time horizons. Both, the stabilization of the column generation and different time horizons, are not subject to our study. They emphasize on the methodological contribution. We combine a theoretical and practical contribution and apply the standard column generation approach, extended by heuristic approaches, to solve the LP-relaxation. Moreover, the underlying graphs differ. As we model ride-sharing directly into the graph we have multiple possibilities to cover tasks and trips. Therefore, we have to make sure that a user is not driving in parallel. Moreover, we allow the vehicle routes to start and end in different depots, which is not the case in \citet{Oukil2007}. 
Similar to \citet{Desfontaines2018}, we work on a multi-graph in which copies of links represent the same connection at different times and, in our case also involving different ride-sharing activities.

\section{Problem description}\label{sec:probform}

We study the multimodal car- and ride-sharing problem in a company having one or more offices from where the employees have to visit various customers during office hours (e.g., for business meetings). Assuming that a company operates different offices {(also referred to as depots)}, an employee might work in any depot. We note that even though the cases where employees switch their work place are rather rare, we included them because they were mentioned by our company partners. Therefore, it is not necessary that the employee returns to the starting depot after her meeting with a customer. Each customer visit involves one specific employee. We consider fixed and unique employee-to-meeting assignments and a fixed latest arrival time. As we consider business meetings at the customer locations, we assume that even if the employee arrives earlier, the starting time of the meetings will not change. Knowing the fixed starting time of the meeting as well as the length of it, we can calculate in advance the earliest departure time of each ride and do not have to explicitly consider the time of the meeting. This results in a fixed sequence of tasks for every employee with several stops, starting and ending at predefined (but possibly different) depots. We call such a fixed sequence of nodes a trip. 

The company operates a finite number of vehicles at each depot and provides possibilities to use other modes of transport such as public transport, bikes, taxis or walk. We assume no start-up cost is associated with vehicles, and depots must have a specific number of vehicles at the beginning and end of the day. With this we assume that we do not have to account for relocations of cars. The employee specifies which modes can be used, since e.g., a person without a driving license cannot be the driver of a car. Moreover, the cars are only interchanged at the depots. This restriction is given from the project partners as changing cars at customer locations would imply too much inconvenience and they reported limited acceptance for handing over cars during a trip. For example, as the meetings of different users are usually not at the same location and/or time, one would need additional {meeting} points and/or times for the hand-over of the car. Further we consider ride-sharing, which is allowed between users, even when two rides do not have the same origin and/or destination. Ride-sharing may at most involve one co-rider. For further details on users, trips and MOTs see Section~\ref{sec:userstripsmots}. Ride-sharing is described in more detail in Section~\ref{sec:ride-sharing}.

The MMCRP aims at determining the optimal MOT-assignment for each trip and to schedule the routes of the cars, maximizing savings when using a car including ride-sharing compared to any other mobility type whilst ensuring that all customers are visited at the right time by the right employee. The cost for the savings calculations does not only include distance cost but also cost of time (as hourly wages of employees) in order to properly reflect the trade-off between fast (but expensive) and slow (but cheap) MOTs. The savings {calculation} is outlined in Section~\ref{sec:savings}. A vehicle route depicts a route of a vehicle during the day encompassing one or more drivers, handing over the vehicle at a depot including possible ride-sharing activities. Note that for our problem it is sufficient to only explicitly model car routes, as we only schedule the trips for the limited resource (i.e.{,} cars). The remaining trips are assigned to the cheapest other MOT a user is willing to choose. This relies on the realistic assumption that users will rationally choose the next cheapest possibility to travel, if a car is not available. For a better understanding of the problem, an illustrative example is given in Section~\ref{sec:illustrative}. 

\subsection{Users, trips and modes of transport}\label{sec:userstripsmots}

We have $u$ users and $n$ tasks given. Each user $p$ has a sequence of tasks $Q_p = (q^1_p, q^2_p, \ldots, q^{n_p}_p)$ that need to be covered. User $p$ starts
at depot $a_p$ and finishes at a (possibly different) depot $b_p$ according to the user's wishes. A trip $\pi$ denotes the sequence of nodes of user $p$ starting at $a_p$ and ending at $b_p$.
Each task $q^i_p$ is associated with a latest arrival time and earliest departure time{. A}s we assume a fixed starting time of the task (i.e., the latest arrival time) we also know the earliest departure time by adding the duration of the task to it. Doing so, we do not have to explicitly consider the duration. 
The driving time between two tasks $(q^i_p,q^j_p)$ using mode of transport $k$ is $t_{q^i_p,q^j_p}^k$, while the cost is $c_{q^i_p,q^j_p}^k$. 
We consider a set of modes of transport $K = \{car,walk,bike,public,taxi\}$. 
Every user $p$ has a set $K^p \subseteq K$ of possible modes of transport that can be used. 
We assume that a pool $W_d$ of shared cars is available at depot $d \in D$ at the beginning of the day, 
and that $\overline{W}_d$ cars should be returned to the depot at the end of the day. Depots $d \in D$ reflect the depots where the cars are parked and the trips start and end. Note that all start and end nodes of a trip $a_p,b_p$ are connected to the depots $d$. The demand at the end of a day will typically reflect the forecasted cars needed at the depot on the following day.
For the other mobility types (car, walk, bike, taxi), we assume that there is infinite capacity.

If a trip $\pi$ is started by a car, then the car should be used for the full trip. 
However, ride-sharing can take place between any two {nodes of a trip} driven by the car. If the co-riding user does not follow the driver for the full trip, then we assume that the cheapest other MOT is used for the rest of the trip. We assume that if a user does not use a car on her own or is not co-riding, then she will take the cheapest other MOT included in her set of MOTs in order to conduct her trip. 

\noindent
\subsection{Ride-sharing}\label{sec:ride-sharing}
Employees can share a ride if it is beneficial. {Usually t}his applies if meetings are visited together or different meetings are nearby or lie on the colleague's trip.
We distinguish between three ride-sharing types: (1) co-riding users share the same origin and destination, (2) they have the same origin and distinct destinations or vice versa, (3) they have different origins and destinations. In the following some representative examples are provided. 

Each user $p$ has to cover a set of tasks $Q_p = (q^1_p, q^2_p, \ldots, q^{n_p}_p)$.
We are considering ride-sharing between two users $p=1$ and $p=2$ on a leg between two tasks {($q^i_p,q^j_p$)}, thus going from $q^1_1$ to $q^2_1$ for user $p=1$ and another leg going from $q^1_2$ to $q^2_2$ for user $p=2$. Although our framework can easily be generalized to multi-user ride-sharing, we only consider two user ride-sharing to ensure user satisfaction. By allowing multiple users to share a ride, a user might end up using a disproportional part of the time as a driver for others.

\begin{figure}
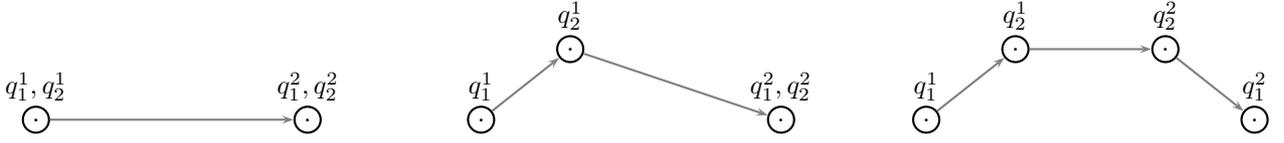

\subcaptionbox{Two rides with same origin and destination in time-space representation.}
[0.33\textwidth]
{
$$
\mbox{\small$
\psmatrix[mnode=circle,colsep=8mm,rowsep=3mm]
 . &   &   &   & . \\
\endpsmatrix
\psset{shortput=nab,arrows=->,labelsep=2pt}
\ncarc[linecolor=gray,arcangle=00]{1,1}{1,5}_{{\red }{\blue }}
\nput{90}{1,1}{q^1_1,q^1_2}
\nput{90}{1,5}{q^2_1,q^2_2}
$}\medskip
$$
}
\subcaptionbox{Two rides with shared destination in time-space representation.}
[0.33\textwidth]
{
$$
\mbox{\small$
\psmatrix[mnode=circle,colsep=8mm,rowsep=3mm]
   & .             \\
 . &   &   &   & . \\
\endpsmatrix
\psset{shortput=nab,arrows=->,labelsep=2pt}
\ncarc[linecolor=gray,arcangle=00]{2,1}{1,2}_{{\red }{\blue }}
\ncarc[linecolor=gray,arcangle=00]{1,2}{2,5}^{{\red }{\blue }}
\nput{90}{1,2}{q^1_2}
\nput{90}{2,1}{q^1_1}
\nput{90}{2,5}{q^2_1,q^2_2}
$}\medskip
$$
}
\subcaptionbox{Two rides with distinct origin and destination in time-space representation.}
[0.33\textwidth]
{
$$
\mbox{\small$
\psmatrix[mnode=circle,colsep=8mm,rowsep=3mm]
   & . &   & .     \\
 . &   &   &   & . \\
\endpsmatrix
\psset{shortput=nab,arrows=->,labelsep=2pt}
\ncarc[linecolor=gray,arcangle=00]{2,1}{1,2}_{{\red }{\blue }}
\ncarc[linecolor=gray,arcangle=00]{1,2}{1,4}^{{\red }{\blue }}
\ncarc[linecolor=gray,arcangle=00]{1,4}{2,5}_{{\red }{\blue }}
\nput{90}{1,2}{q^1_2}
\nput{90}{1,4}{q^2_2}
\nput{90}{2,1}{q^1_1}
\nput{90}{2,5}{q^2_1}
$}\medskip
$$
}
%
\caption{Three examples of ride-sharing illustrated in a time-space network. User $p=1$ is going
from $q^1_1$ to $q^2_1$ while user $p=2$ is going from $q^1_2$ to $q^2_2$.} \label{fig:share}
\end{figure}

The simplest ride-sharing case occurs when users $p=1$ and $p=2$ have the same origin and destination, as shown in Figure \ref{fig:share}(a). In this case both users can be served on the ride.

We can also have the case in which only the source or destination is shared. Starting with the case where the end destination is shared we have the case shown in Figure \ref{fig:share}(b). In this case, user $p=1$ has to drive from $q^1_1$ to $q^1_2$ to pick up user $p=2$, and then both drive to the shared end destination where $q^2_1$ = $q^2_2$. Provided that all time limits are satisfied, the cost of the ride can be calculated as the sum of the individual legs. The case where the origin is shared is handled in a symmetric way. Obviously, this shared ride is only beneficial if the detour for $p=1$ is not too large.

In general, both origin and destination can be distinct as illustrated in Figure \ref{fig:share}(c). In this case user $p=1$ has to drive from $q^1_1$ to $q^1_2$ to pick up user $p=2$, drive this user to her destination $q^2_2$ and then drive to her own destination $q^2_1$. Provided that all time limits are satisfied, the cost of the ride can be calculated as the sum of the individual legs. Please note that the end destination of the driver ($q^2_1$) must always lie after an intermediate point (e.g., $q^2_2$) as we do not allow to change drivers on the trip.

\noindent
\subsection{Savings calculation}\label{sec:savings}

We calculate cost and travel time for each trip. In order to reach the best choice of MOT combination we aim to obtain savings when using a car including ride-sharing compared to any other mobility type rather than focusing on minimizing costs only.  Saving $\gamma_\pi$ of a trip $\pi$ is the sum over all savings $\gamma_{q^i_p}$ of included tasks $q$ on trip $\pi$. We consider costs of subsequent tasks $(q^i_p,q^j_p)$  and obtain the savings calculation by considering detouring for ride-sharing. Note that the obtained savings might also be negative. This will occur if the cheapest MOT for a trip is not the car.

In a first step let us compute the savings obtained when no ride-sharing between two subsequent tasks $(q^i_p,q^j_p)$ is considered. The savings of tasks $q^i_p$ and its fixed successor $q^j_p$ can be calculated as the difference between cost of using the cheapest other MOT $c^{k}_{(q^i_p,q^j_p)}$ and cost of using the car $c^{car}_{(q^i_p,q^j_p)}$, such that:
\begin{equation}
\gamma_{q^i_p} = \min_{k \in K \setminus \{ car \} } \{ c^{k}_{(q^i_p,q^j_p)} - c^{car}_{(q^i_p,q^j_p)}\}
\label{eq:savings1}
\end{equation}
Next, let us assume two users $p=1$ and $p=2$ whereas ride-sharing is demanded between two tasks $(i,j)$ of user $p=2$, $(q^i_2,q^j_2)$. We add detour costs to go to/from these tasks. We have to account for additional costs of going from the driver's {(}$p=1${)} task $q^i_1$ to the starting point of the demanded ride-sharing $q^i_2$ as well as additional cost from the ride-sharing drop-off point $q^j_2$ to the driver's original task $q^j_1$. This gives us an additional detouring cost between $(q^i_1,q^i_2)$ as well as $(q^j_2,q^j_1)$ for which we only take into account the cost of using the car. 
We do not change the fixed sequence of a user's trip, however, we must keep track of whether ride-sharing is conducted between two trips in order to take into account additional detouring cost. 
Hence for not traversing the original link $(q^i_1,q^j_1)$ we save $c^k_{(q^i_1,q^j_1)}$ and additionally save costs by allowing for ride-sharing between $c^k_{(q^i_2,q^j_2)}$. Therefore, for each task $q^i_1$ including subsequent ride-sharing we compute the savings $\gamma_{q^i_1}$ as follows:
\begin{equation}
\gamma_{q^i_1} = \min_{k \in K \setminus \{ car \} } \{ (c^{k}_{(q^i_1,q^j_1)} + c^{k}_{(q^i_2,q^j_2)}) - (c^{car}_{(q^i_2,q^j_2)} + c^{car}_{(q^i_1,q^i_2)}+ c^{car}_{(q^j_2,q^j_1)})\}
\label{eq:savings2}
\end{equation}
assuming $q^j_1$ is its fixed successor and ride-sharing is employed for tasks $(q^i_2,q^j_2)$ between the original sequence $(q^i_1,q^j_1)$.
The saving of trip $\pi$ is then calculated as 
$\gamma_{\pi} = \sum_{q \in \pi} \gamma_q$.

In order to provide a more understandable overview, let us assume a trip $\pi$ of user $p=1$ considering three tasks to be visited, $q^1_1$, $q^2_1$ and $q^3_1$. The trip starts at $a_1$ and ends at $b_1$. The user's trip would then be $a_1$ - $q^1_1$ - $q^2_1$ - $q^3_1$ - $b_1$. The corresponding savings calculation for trip $\pi$ and without considering ride-sharing is as follows: 
$\gamma_{\pi}$ = $\min_{k \in K \setminus \{ car \} } \{$($c^{k}_{(a_1,q^1_1)}$ - $c^{car}_{(a_1,q^1_1)}$) + 
($c^{k}_{(q^1_1,q^2_1)}$ - $c^{car}_{(q^1_1,q^2_1)}$) + 
($c^{k}_{(q^2_1,q^3_1)}$ - $c^{car}_{(q^2_1,q^3_1)}$) + 
($c^{k}_{(q^3_1,b_1)}$ - $c^{car}_{(q^3_1,b_1)}$)$\}$.  
In a next step, assume that user $p=1$ takes a detour between $q^2_1$ and $q^3_1$ in order to ride-share with user $p=2$ between $q^1_{2}$ and $q^2_{2}$. Now the sequence would be $a_1$ - $q^1_1$ - $q^2_1$ - $q^1_{2}$ - $q^2_{2}$ - $q^3_1$ - $b_1$ whereas the fixed successors for our calculations do not change, as described above. We get the respective savings of the above presented trip $\pi$:
$\gamma_{\pi}$ = $\min_{k \in K \setminus \{ car \} } \{$($c^{k}_{(a_1,q^1_1)}$ - $c^{car}_{(a_1,q^1_1)}$) + 
($c^{k}_{(q^1_1,q^2_1)}$ - $c^{car}_{(q^1_1,q^2_1)}$) + 
($c^{k}_{(q^2_1,q^3_1)}$ + $c^{k}_{(q^1_{2},q^2_{2})}$ - $c^{car}_{(q^1_{2},q^2_{2})}$ -  $c^{car}_{(q^2_1,q^1_{2})}$ - $c^{car}_{(q^2_{2},q^3_1)}$) +
($c^{k}_{(q^3_1,b_1)}$ - $c^{car}_{(q^3_1,b_1)}$)$\}$.
We make these calculations for every variant of trip $\pi$ representing all possible ride-sharing trips.

Please note that the same task $q^i_p$ can be on different trips and have different savings as they represent different rides. Moreover, for the cases where the driver $p$ and rider $p'$ share their origin and/or destination we do not account for all detouring cost such that $c^{car}_{(q^i_p,q^i_{p'})} = 0$ and/or  $c^{car}_{(q^j_{p'},q^j_p)} = 0$, respectively. 

\subsection{Illustrative example}\label{sec:illustrative}

\begin{figure}
\begin{center}
\begin{subfigure}{0.475\textwidth}\centering
\begin{tikzpicture}[scale=0.4]
\draw[gray, ->] (-3,-2) -- (3,-2) node [midway, above, sloped] (TextNode) {\tiny time};
\draw[gray, ->] (-3,-2) -- (-3,3) node [midway, above, sloped] (TextNode) {\tiny user's schedule};
\tikzstyle{every node}=[draw,shape=circle];
\tiny
\path (-2,0)  node[draw=white] (person) {$p=4$:};
\path (-2,2.5)  node[draw=white] (person) {$p=3$:};
\path (-2,5)  node[draw=white] (person) {$p=2$:};
\path (-2,7.5)  node[draw=white] (person) {$p=1$:};
\begin{scope}[on background layer]
\draw[pattern=horizontal lines] (0.3,8.2) rectangle (1.9,6.8);
\draw[preaction={fill=purple},pattern=dots] (1.9,8.2) rectangle (4.3,6.8);
\draw[pattern=horizontal lines] (4.3,8.2) rectangle (5.7,6.8);
\draw[preaction={fill=purple},pattern=dots] (5.7,8.2) rectangle (6.3,6.8);
\draw[pattern=horizontal lines] (6.3,8.2) rectangle (7.7,6.8);
\draw[preaction={fill=purple},pattern=dots] (7.7,8.2) rectangle (13.9,6.8);
\draw[pattern=horizontal lines] (13.9,8.2) rectangle (15.3,6.8);
\draw[preaction={fill=purple},pattern=dots] (15.3,8.2) rectangle (17.7,6.8);
\draw[pattern=horizontal lines] (17.7,8.2) rectangle (19.2,6.8);
\draw[pattern=horizontal lines] (-1,5.7) rectangle (1,4.3);
\draw[pattern=dots] (1,5.7) rectangle (2.3,4.3);
\draw[pattern=horizontal lines] (2.3,5.7) rectangle (3.7,4.3);
\draw[pattern=dots] (3.7,5.7) rectangle (9.2,4.3);
\draw[pattern=horizontal lines] (9.2,5.7) rectangle (10.7,4.3);
\draw[pattern=horizontal lines] (1,3.2) rectangle (3,1.8);
\draw[pattern=dots] (3,3.2) rectangle (7.7,1.8);
\draw[pattern=horizontal lines] (7.7,3.2) rectangle (9.1,1.8);
\draw[pattern=dots] (9.1,3.2) rectangle (11.2,1.8);
\draw[pattern=horizontal lines] (11.2,3.2) rectangle (12.7,1.8);
\draw[pattern=dots] (12.7,3.2) rectangle (13.2,1.8);
\draw[pattern=horizontal lines] (13.2,3.2) rectangle (14.6,1.8);
\draw[pattern=dots] (14.6,3.2) rectangle (17.7,1.8);
\draw[pattern=horizontal lines] (17.7,3.2) rectangle (19.2,1.8);
\draw[pattern=horizontal lines] (0.8,0.7) rectangle (2.3,-0.7);
\draw[preaction={fill=purple},pattern=dots] (2.3,0.7) rectangle (2.8,-0.7);
\draw[pattern=horizontal lines] (2.8,0.7) rectangle (4.3,-0.7);
\draw[preaction={fill=purple},pattern=dots] (4.3,0.7) rectangle (10.4,-0.7);
\draw[pattern=horizontal lines] (10.4,0.7) rectangle (11.9,-0.7);
\draw[preaction={fill=purple},pattern=dots] (11.9,0.7) rectangle (16.1,-0.7);
\draw[pattern=horizontal lines] (16.1,0.7) rectangle (17.7,-0.7);
\end{scope}
\draw[->, very thick] (1.9,7.5) to (4.3,7.5);
\draw[->, very thick]  (5.7,7.5) to (6.3,7.5);
\draw[->, very thick]  (7.7,7.5) to (13.9,7.5);
\draw[->, very thick] (15.3,7.5) to (17.7,7.5);
\draw[->, very thick] (2.3,0) to (2.8,0);
\draw[->, very thick] (4.3,0) to (10.4,0);
\draw[->, very thick] (11.9,0) to (16.1,0);
\path (1.5,0)  node (depot) [fill=white,shape=diamond] {$d_1$};
\path (2,2.5)  node (depot1) [fill=white,shape=diamond] {$d_2$};
\path (1,7.5)  node (depot2) [fill=white,shape=diamond] {$d_1$};
\path (0,5)  node (depot3) [fill=white,shape=diamond] {$d_2$};
\path (3.5,0)    node (1) [fill=white] {$q^1_4$};
\path (12,2.5) node (depot4) [fill=white,shape=diamond] {$d_1$};
\path (7,7.5) node (7) [fill=white] {$q^2_1$};
\path (5,7.5) node (8) [fill=white] {$q^1_1$};
\path (10,5) node (depot5) [fill=white,shape=diamond] {$d_1$};
\path (3,5) node (4) [fill=white] {$q^1_2$};
\path (8.5,2.5) node (5) [fill=white] {$q^1_3$};
\path (11,0) node (6) [fill=white] {$q^2_4$};
\path (17,0) node (depot7) [fill=white,shape=diamond] {$d_1$};
\path (14,2.5) node (2) [fill=white] {$q^2_3$};
\path (18.5,2.5) node (depot8) [fill=white,shape=diamond] {$d_2$};
\path (18.5,7.5) node (depot9) [fill=white,shape=diamond] {$d_2$};
\path (14.5,7.5) node (9) [fill=white] {$q^3_1$};
\end{tikzpicture}
\caption{Example solution without sharing.}
\end{subfigure} \hfill
\begin{subfigure}{0.475\textwidth}\centering
\begin{tikzpicture}[scale=0.4]
\draw[white, ->] (-3,-2) -- (3,-2) node [midway, above, sloped] (TextNode) {\textcolor{white}{time}};
\draw[white, ->] (-3,-2) -- (-3,3) node [midway, above, sloped] (TextNode) {\textcolor{white}{user}};
\tikzstyle{every node}=[draw,shape=circle];
\tiny
\path (-2,0)  node[draw=white] (person) {$p=4$:};
\path (-2,2.5)  node[draw=white] (person) {$p=3$:};
\path (-2,5)  node[draw=white] (person) {$p=2$:};
\path (-2,7.5)  node[draw=white] (person) {$p=1$:};
\begin{scope}[on background layer]
\draw[pattern=horizontal lines] (0.3,8.2) rectangle (1.9,6.8);
\draw[pattern=dots] (1.9,8.2) rectangle (4.3,6.8);
\draw[pattern=horizontal lines] (4.3,8.2) rectangle (5.7,6.8);
\draw[preaction={fill=yellow},pattern=dots] (5.7,8.2) rectangle (6.3,6.8);
\draw[pattern=horizontal lines] (6.3,8.2) rectangle (7.7,6.8);
\draw[pattern=dots] (7.7,8.2) rectangle (13.9,6.8);
\draw[pattern=horizontal lines] (13.9,8.2) rectangle (15.3,6.8);
\draw[preaction={fill=yellow},pattern=dots] (15.3,8.2) rectangle (17.7,6.8);
\draw[pattern=horizontal lines] (17.7,8.2) rectangle (19.2,6.8);
\draw[pattern=horizontal lines] (-1,5.7) rectangle (1,4.3);
\draw[preaction={fill=purple},pattern=dots] (1,5.7) rectangle (2.3,4.3);
\draw[pattern=horizontal lines] (2.3,5.7) rectangle (3.7,4.3);
\draw[preaction={fill=purple},pattern=dots] (3.7,5.7) rectangle (5.7,4.3);
\draw[preaction={fill=yellow},pattern=dots] (5.7,5.7) rectangle (6.3,4.3);
\draw[preaction={fill=purple},pattern=dots] (6.3,5.7) rectangle (9.2,4.3);
\draw[pattern=horizontal lines] (9.2,5.7) rectangle (10.7,4.3);
\draw[pattern=horizontal lines] (1,3.2) rectangle (3,1.8);
\draw[pattern=dots] (3,3.2) rectangle (7.7,1.8);
\draw[pattern=horizontal lines] (7.7,3.2) rectangle (9.1,1.8);
\draw[pattern=dots] (9.1,3.2) rectangle (11.2,1.8);
\draw[pattern=horizontal lines] (11.2,3.2) rectangle (12.7,1.8);
\draw[preaction={fill=purple},pattern=dots] (12.7,3.2) rectangle (13.2,1.8);
\draw[pattern=horizontal lines] (13.2,3.2) rectangle (14.6,1.8);
\draw[preaction={fill=purple},pattern=dots] (14.6,3.2) rectangle (15.3,1.8);
\draw[preaction={fill=yellow},pattern=dots] (15.3,3.2) rectangle (17.7,1.8);
\draw[pattern=horizontal lines] (17.7,3.2) rectangle (19.2,1.8);
\draw[pattern=horizontal lines] (0.8,0.7) rectangle (2.3,-0.7);
\draw[preaction={fill=purple},pattern=dots] (2.3,0.7) rectangle (2.8,-0.7);
\draw[pattern=horizontal lines] (2.8,0.7) rectangle (4.3,-0.7);
\draw[preaction={fill=purple},pattern=dots] (4.3,0.7) rectangle (10.4,-0.7);
\draw[pattern=horizontal lines] (10.4,0.7) rectangle (11.9,-0.7);
\draw[preaction={fill=purple},pattern=dots] (11.9,0.7) rectangle (16.1,-0.7);
\draw[pattern=horizontal lines] (16.1,0.7) rectangle (17.7,-0.7);
\end{scope}
\draw[->, very thick]  (5.7,7.5) to (6.3,7.5);
\draw[->, very thick] (15.3,7.5) to (17.7,7.5);
\draw[->, very thick] (1,5) to (2.3,5);
\draw[->, very thick] (3.7,5) to (5.7,6.8);
\draw[->, very thick] (6.3,6.8) to (9.2,5);
\draw[->, very thick] (12.7,2.5) to (13.2,2.5);
\draw[->, very thick] (10.7,4.3) to (11.2,3.2);
\draw[->, very thick] (14.6,2.5) to (15.3, 6.8);
\draw[->, very thick] (17.7,6.8) to (17.7,2.5);
\draw[->, very thick] (2.3,0) to (2.8,0);
\draw[->, very thick] (4.3,0) to (10.4,0);
\draw[->, very thick] (11.9,0) to (17.7,0);
\path (1.5,0)  node (depot) [fill=white,shape=diamond] {$d_1$};
\path (2,2.5)  node (depot1) [fill=white,shape=diamond] {$d_2$};
\path (1,7.5)  node (depot2) [fill=white,shape=diamond] {$d_1$};
\path (0,5)  node (depot3) [fill=white,shape=diamond] {$d_2$};
\path (3.5,0)    node (1) [fill=white] {$q^1_4$};
\path (12,2.5) node (depot4) [fill=white,shape=diamond] {$d_1$};
\path (7,7.5) node (7) [fill=white] {$q^2_1$};
\path (5,7.5) node (8) [fill=white] {$q^1_1$};
\path (10,5) node (depot5) [fill=white,shape=diamond] {$d_1$};
\path (3,5) node (4) [fill=white] {$q^1_2$};
\path (8.5,2.5) node (5) [fill=white] {$q^1_3$};
\path (11,0) node (6) [fill=white] {$q^2_4$};
\path (17,0) node (depot7) [fill=white,shape=diamond] {$d_1$};
\path (14,2.5) node (2) [fill=white] {$q^2_3$};
\path (18.5,2.5) node (depot8) [fill=white,shape=diamond] {$d_2$};
\path (18.5,7.5) node (depot9) [fill=white,shape=diamond] {$d_2$};
\path (14.5,7.5) node (9) [fill=white] {$q^3_1$};
\end{tikzpicture}
\caption{Example solution with car- and ride-sharing.}
\end{subfigure}
\begin{tikzpicture}
\scriptsize
\draw [->, thick] (0,0) -- (1,0) node [midway, above, sloped] (TextNode) {car};
\path (2,0) node (node_) [draw,shape=circle] {task};
\path (4,0) node (node_) [draw,shape=diamond] {depot};
\path (6,0) node (node_) [draw,shape=rectangle,pattern=horizontal lines] {work};
\path (8,0) node (node_) [draw,shape=rectangle,pattern=dots] {travel};
\path (10,0) node (node_) [draw,shape=rectangle,fill=yellow] {ride-share};
\path (12,0) node (node_) [draw,shape=rectangle,fill=purple] {car};
\path (14,0) node (node_) [draw,shape=rectangle,fill=white] {cheapest other MOT};
\end{tikzpicture}
\end{center}
\small
\renewcommand{\baselinestretch}{1.0}
\caption{Examples without sharing as well as car- and ride-sharing. We have two offices (depots) $d_1$ and $d_2$ and four users $p=1$, $p=2$, $p=3$ and $p=4$; tasks are denoted as $q^i_p$. Background rectangles with lines depict duration of a meeting, dots mean the user is traveling. If the background is not colored, the user is traveling with the cheapest other MOT, purple depicts travel by car, yellow ride-sharing. The arrows illustrate the routes of the two cars. } \label{fig:example}
\end{figure}
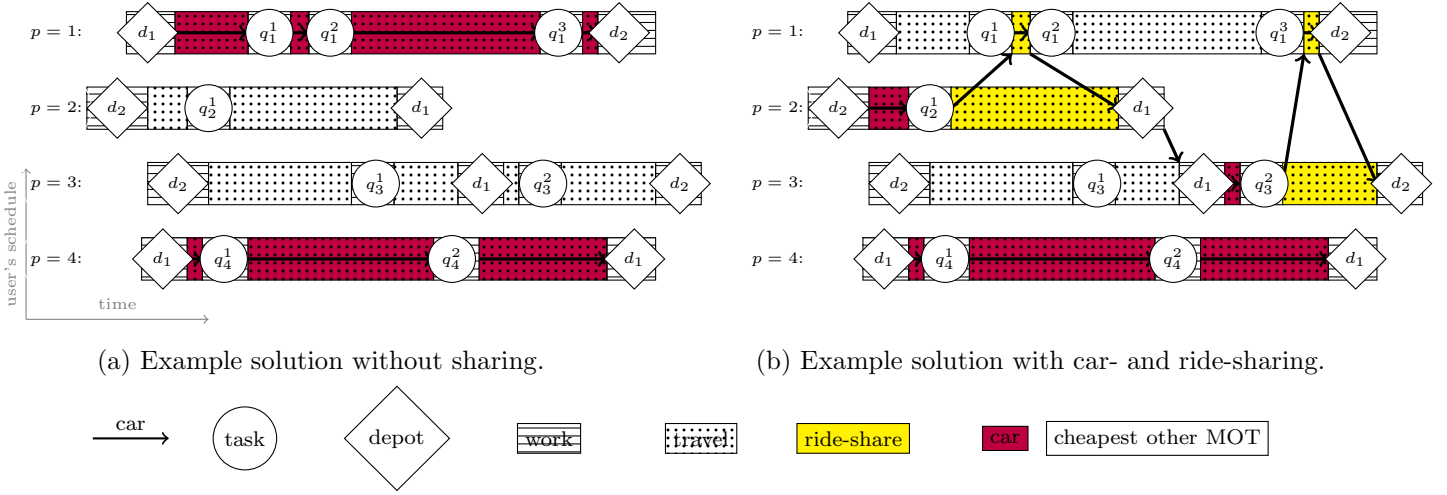

To better illustrate the problem, a possible schedule is shown in Figure \ref{fig:example}.
We have 4 users ($p=1$,$p=2$,$p=3$,$p=4$), 2 cars, and 2 depots ($d_1$ and $d_2$). Each user's schedule is depicted by one horizontal line connecting depots $d$ and meetings $q^i_p$. User $p=1$ visits $q^1_1$,$q^2_1$ and $q^3_1$, user $p=2$ is assigned to $q^1_2$, user $p=3$ visits $q^1_3$ and $q^2_3$ whilst returning in between to depot $d_1$ and lastly user $p=4$ drives to tasks $q^1_4$ and $q^2_4$. Background rectangles with lines depict duration of a meeting, dots indicate the user is traveling. If the background is not colored, the user is traveling with the cheapest other MOT, purple denotes travel by car, yellow ride-sharing. The arrows illustrate the traveling of the two cars. Figure~\ref{fig:example}(a) shows a possible solution without sharing, Figure~\ref{fig:example}(b) gives an adapted solution with car- and ride-sharing. User $p=4$ uses in both figures one car for the whole trip and does not share any rides. In Figure~\ref{fig:example}(a) the second car is used by user $p=1$ for the whole trip. Differently in Figure~\ref{fig:example}(b) {where} one of the cars is handed over from user $p=2$ to user $p=3$ at depot $d_1$. Furthermore, both drivers of the car take on user $p=1$ for some legs of her required trip, shown in yellow. Otherwise, user $p=1$ uses the cheapest other MOT. 
\bigskip

\section{Solution approach}
\label{sec:solutionapproach}

In order to simplify the problem, we reformulate it using a time-space network, in which all possible ride-sharing trips are represented by arcs. 
We introduce a graph $G(V,E)$, where vertices are given by their time-space coordinates 
(i.e.{,} time and depot).

Figures~\ref{fig:graphs}(a) and (b) show the first step of the graph transformation. We start by modeling the depots $d$ at the start and end of the planning horizon where all trips are connected to and cars are located, as well as $a_p$ and $b_p$ denoting the start and end points of a trip $\pi$, and user tasks $q \in Q_p$ depicted as $q_p^i$. Solid lines denote trips, dotted ones indicate waiting arcs in the set $E'$ denoting that the car is not moving. Hence, we start by considering all trips of all users starting at $a_p$ and ending at $b_p$ including tasks $q$ through the whole planning horizon, which then start and end at depot $d$. 
In Figure~\ref{fig:graphs}(a) we show the starting point of the graph construction whereas user tasks $q \in Q_p$ are still in the graph. From this graph, we transform all $a_p$, $b_p$ connections (i.e.{,} all trips) to the arcs as represented in Figure~\ref{fig:graphs}(b). As can be seen, we do not explicitly consider the tasks in the graph any more but save all relevant information on the arcs.
For each user $p \in P$ we enumerate all possible trips from the user's start depot $a_p$ to 
the user's end depot $b_p$, including possible ride-sharing as described in Section~\ref{sec:probform}. Figure~\ref{fig:graphs}(c) gives two possible ride-sharing trips. Figure~\ref{fig:graphs}(d) then shows the extension in the trip-based arc representation of Figure~\ref{fig:graphs}(a). As can be seen, this gives us multiple arcs between nodes $a_p$ and $b_p$ each depicting a possible way how the trip can be conducted.
Due to ride-sharing, the start time of two rides may be different even if they consider the
same user $p$ starting at the same depot $a$. A similar observation holds for the end times
of two rides for user $p$.

Every possible trip ${\pi} \in R_p$ of user $p$ including any number of co-ride possibilities {(}including 0{)} results in a tuple
$\{(a_p^{\pi}, b_p^{\pi}, g_p^{\pi}, h_p^{\pi}, s_p^{\pi}, \ell_p^{\pi})\}$.
Here we have that
$a_p^{\pi}$ is the start depot of the ride trip $\pi$,
$b_p^{\pi}$ is the end depot of trip $\pi$,
$g_p^{\pi}$ is the departure time at start depot $a_p^{\pi}$,
$h_p^{\pi}$ is the arrival time at end depot $b_p^{\pi}$,
$s_p^{\pi}$ is the saving of the ride,
and $\ell_p^{\pi}$ is a list of visits covered, both for the driver and possible co-rider(s).

For every user $p$ 
let $V_p = \{ (a_p^{\pi}, g_p^{\pi}) \}_{{\pi} \in R_p} \cup \{ (b_p^{\pi}, h_p^{\pi}) \}_{{\pi} \in R_p}$ 
be the set of nodes associated with $p$ as driver of the car.
Let the set of ride arcs be $E_p = \{ \left( (a_p^{\pi}, g_p^{\pi}) , (b_p^{\pi}, h_p^{\pi}) \right) \}_{{\pi} \in R_p}$. 
The saving of a ride arc $s_p^{\pi}$.

If we assume that $\sigma$ is the first possible time, and $\tau$ is the last possible 
time in the planning horizon, then for every depot we construct nodes $(d,\sigma)$ and
$(d,\tau)$. The set of nodes can now be defined as
$$
  V = \{V_p\}_{p\in P} \cup \{(d,\sigma)\}_{d\in D} \cup \{(d,\tau)\}_{d\in D}
$$
Finally we need to introduce the set $E'$ of waiting arcs. A waiting arc is
inserted between nodes $(d,h'), (d,h'') \in V$ in the graph if they correspond to 
the same depot $d$ and $h''$ comes immediately after $h'$ 
(i.e.{,} no other node $(d,h)$ with $h' < h < h''$ exist).
Now, the set of arcs can be defined as
$$
  E = \{E_p\}_{p\in P} \cup E' 
$$
Due to the possibly exponential number of ride-sharing combinations, we can have an 
exponential number of arcs. However, in practice, the number of possible trips ${\pi}$ for each
user $p$ is quite limited.

\begin{figure}[!ht]
\renewcommand{\baselinestretch}{1.0}
\subcaptionbox{
First step of the graph construction: graph including information on tasks.}[0.4\textwidth]
{%
$$
\mbox{\small$
\psscalebox{0.7}{
\psmatrix[mnode=circle,colsep=4mm,rowsep=5mm]
			&       &		&		&		&   	&q^2_2									\\
			&       &		&		&		&q^1_2									\\
   d_1	&a_2       & a_3	&	&b_3	&a_1	&		&		&b_2	&		& d_1    \\
   			&       &		&q^1_3	&		&		&q^1_1									\\
   			&       &		&		&q^1_5	&		&		&q^2_1						\\
   d_2	&       & 		&a_5	&a_4		&b_5	&	&		&b_1	&b_4  	& d_2  \\
   			&       &		&		&		&		&		&       &q^1_4							\\	
\endpsmatrix
\psset{shortput=nab,arrows=->,labelsep=2pt}
\ncarc[linestyle=dotted,arcangle=00]{3,1}{3,2}
\ncarc[linestyle=dotted,arcangle=00]{3,2}{3,3}
\ncarc[linestyle=dotted,arcangle=00]{3,3}{3,5}
\ncarc[linestyle=dotted,arcangle=00]{3,5}{3,6}
\ncarc[linestyle=dotted,arcangle=00]{3,6}{3,9}
\ncarc[linestyle=dotted,arcangle=00]{3,9}{3,11}
\ncarc[linestyle=dotted,arcangle=00]{6,1}{6,4}
\ncarc[linestyle=dotted,arcangle=00]{6,4}{6,5}
\ncarc[linestyle=dotted,arcangle=00]{6,5}{6,6}
\ncarc[linestyle=dotted,arcangle=00]{6,6}{6,9}
\ncarc[linestyle=dotted,arcangle=00]{6,9}{6,10}
\ncarc[linestyle=dotted,arcangle=00]{6,10}{6,11}
\ncarc[arcangle=00]{3,2}{2,6}
\ncarc[arcangle=00]{2,6}{1,7}
\ncarc[arcangle=00]{1,7}{3,9}
\ncarc[arcangle=00]{3,3}{4,4}
\ncarc[arcangle=00]{4,4}{3,5}
\ncarc[arcangle=00]{3,6}{4,7}
\ncarc[arcangle=00]{4,7}{5,8}
\ncarc[arcangle=00]{5,8}{6,9}
\ncarc[arcangle=00]{7,4}{6,5}
\ncarc[arcangle=00]{6,5}{7,9}
\ncarc[arcangle=00]{7,9}{6,10}
\ncarc[arcangle=00]{6,4}{5,5}
\ncarc[arcangle=00]{5,5}{6,6}
}
$}\medskip
$$
} 
\hfill
\subcaptionbox{Second step of the graph construction: modeling the trips as arcs.}
[0.4\textwidth]
{
$$
\mbox{\small$
\psscalebox{0.7}{
\psmatrix[mnode=circle,colsep=4mm,rowsep=30mm]
   d_1	&a_2       & a_3	&	&b_3	&a_1	&		&		&b_2	&		& d_1    \\
   d_2	&       & 		&a_5	&a_4		&b_5	&	&		&b_1	&b_4  	& d_2  \\
\endpsmatrix
\psset{shortput=nab,arrows=->,labelsep=2pt}
\ncarc[linestyle=dotted,arcangle=00]{1,1}{1,2}
\ncarc[linestyle=dotted,arcangle=00]{1,2}{1,3}
\ncarc[linestyle=dotted,arcangle=00]{1,3}{1,5}
\ncarc[linestyle=dotted,arcangle=00]{1,5}{1,6}
\ncarc[linestyle=dotted,arcangle=00]{1,6}{1,9}
\ncarc[linestyle=dotted,arcangle=00]{1,9}{1,11}
\ncarc[linestyle=dotted,arcangle=00]{2,1}{2,4}
\ncarc[linestyle=dotted,arcangle=00]{2,4}{2,5}
\ncarc[linestyle=dotted,arcangle=00]{2,5}{2,6}
\ncarc[linestyle=dotted,arcangle=00]{2,6}{2,9}
\ncarc[linestyle=dotted,arcangle=00]{2,9}{2,10}
\ncarc[linestyle=dotted,arcangle=00]{2,10}{2,11}
\ncarc[arcangle=40]{1,2}{1,9}
\ncarc[arcangle=-40]{1,3}{1,5}
\ncarc[arcangle=00]{1,6}{2,9}
\ncarc[arcangle=40]{2,4}{2,6}
\ncarc[arcangle=-40]{2,5}{2,10}
}
$}\medskip
$$
\vspace{.5cm}
} \\
\subcaptionbox{Graph with two ride-sharing trips.}
[0.4\textwidth]
{
$$
\mbox{\small$
\psscalebox{0.7}{
\psmatrix[mnode=circle,colsep=4mm,rowsep=5mm]
			&       &		&		&		&   	&q^2_2									\\
			&       &		&		&		&q^1_2									\\
   d_1	&a_2       & a_3	&	&b_3	&a_1	&		&		&b_2	&		& d_1    \\
   			&       &		&q^1_3	&		&		&q^1_1									\\
   			&       &		&		&q^1_5	&		&		&q^2_1						\\
   d_2	&       & 		&a_5	&a_4		&b_5	&	&		&b_1	&b_4  	& d_2  \\
   			&       &		&		&		&		&		&       &q^1_4							\\	
\endpsmatrix
\psset{shortput=nab,arrows=->,labelsep=2pt}
\ncarc[linestyle=dotted,arcangle=00]{3,1}{3,2}
\ncarc[linestyle=dotted,arcangle=00]{3,2}{3,3}
\ncarc[linestyle=dotted,arcangle=00]{3,3}{3,5}
\ncarc[linestyle=dotted,arcangle=00]{3,5}{3,6}
\ncarc[linestyle=dotted,arcangle=00]{3,6}{3,9}
\ncarc[linestyle=dotted,arcangle=00]{3,9}{3,11}
\ncarc[linestyle=dotted,arcangle=00]{6,1}{6,4}
\ncarc[linestyle=dotted,arcangle=00]{6,4}{6,5}
\ncarc[linestyle=dotted,arcangle=00]{6,5}{6,6}
\ncarc[linestyle=dotted,arcangle=00]{6,6}{6,9}
\ncarc[linestyle=dotted,arcangle=00]{6,9}{6,10}
\ncarc[linestyle=dotted,arcangle=00]{6,10}{6,11}
\ncarc[arcangle=00]{3,2}{3,3}
\ncarc[arcangle=00]{3,3}{4,4}
\ncarc[arcangle=00]{4,4}{2,6}
\ncarc[arcangle=00]{2,6}{1,7}
\ncarc[arcangle=00]{1,7}{3,9}
\ncarc[arcangle=00]{6,5}{4,7}
\ncarc[arcangle=00]{4,7}{5,8}
\ncarc[arcangle=00]{5,8}{7,9}
\ncarc[arcangle=00]{7,9}{6,10}
}
$}\medskip
$$
}\hfill 
\hfill
\subcaptionbox{Extended graph with trips as arcs, where two ride-sharing trips are added to the graph in (b).}
[0.4\textwidth]
{
$$
\mbox{\small$
\psscalebox{0.7}{
\psmatrix[mnode=circle,colsep=4mm,rowsep=30mm]
   d_1	&a_2       & a_3	&	&b_3	&a_1	&		&		&b_2	&		& d_1    \\
   d_2	&       & 		&a_5	&a_4		&b_5	&	&		&b_1	&b_4  	& d_2  \\
\endpsmatrix
\psset{shortput=nab,arrows=->,labelsep=2pt}
\ncarc[linestyle=dotted,arcangle=00]{1,1}{1,2}
\ncarc[linestyle=dotted,arcangle=00]{1,2}{1,3}
\ncarc[linestyle=dotted,arcangle=00]{1,3}{1,5}
\ncarc[linestyle=dotted,arcangle=00]{1,5}{1,6}
\ncarc[linestyle=dotted,arcangle=00]{1,6}{1,9}
\ncarc[linestyle=dotted,arcangle=00]{1,9}{1,11}
\ncarc[linestyle=dotted,arcangle=00]{2,1}{2,4}
\ncarc[linestyle=dotted,arcangle=00]{2,4}{2,5}
\ncarc[linestyle=dotted,arcangle=00]{2,5}{2,6}
\ncarc[linestyle=dotted,arcangle=00]{2,6}{2,9}
\ncarc[linestyle=dotted,arcangle=00]{2,9}{2,10}
\ncarc[linestyle=dotted,arcangle=00]{2,10}{2,11}
\ncarc[arcangle=40]{1,2}{1,9}
\ncarc[arcangle=50]{1,2}{1,9}
\ncarc[arcangle=-40]{1,3}{1,5}
\ncarc[arcangle=00]{1,6}{2,9}
\ncarc[arcangle=40]{2,4}{2,6}
\ncarc[arcangle=-40]{2,5}{2,10}
\ncarc[arcangle=-50]{2,5}{2,10}
}
$}\medskip
$$
\vspace{.5cm}
} \\
\begin{center}
\begin{subfigure}{0.4\textwidth}\centering
\begin{tikzpicture}
\scriptsize
\draw [dotted,->](0,0) -- (1,0) node [midway, above, sloped] (TextNode) {waiting arc};
\draw [->] (2,0) -- (3,0) node [midway, above, sloped] (TextNode) {trip};
\end{tikzpicture}
\end{subfigure}
\end{center}
\caption{Illustration of the auxiliary graph {in a time-space network}.} \label{fig:graphs}
\end{figure}

Note that the problem is modeled as a kind of a vehicle scheduling problem with multiple depots. In the vehicle scheduling problem with multiple depots we have a set $Z$ of trips. A trip $z$ has an associated start time $l_z$ and end time $l_z'$. Two trips $z$ and $z'$ can be run in sequence (i.e.{,} they are compatible) if there is sufficient time to get from $z$ to $z'$. Moreover, we have multiple depots, each depot $d$ having a capacity $W_d$. Every trip has to be run by one vehicle, minimizing the driving costs. {A major} difference between the vehicle scheduling problem and the MMCRP is that in our case not all trips need to be covered. However, the vehicle scheduling problem can either be transformed into the MMCRP by having infinite cost for all other modes of transport, forcing the solution to cover all trips or we it can be seen as a{n extended} vehicle scheduling problem with profits.

\noindent
\textbf{Complexity:} 
As elaborated, the problem is modeled as a kind of a vehicle scheduling problem. Therefore, it is easy to show that the MMCRP is NP-hard if the number of depots is at least two.
We prove the complexity by reduction from the \myemph{vehicle scheduling problem with 2 depots} which was proven to be NP-hard by \citet{Bertossi1987}.
Notice that, since the vehicle scheduling problem does not consider ride-sharing, the MMCRP with at least 2 depots is NP-hard  even without ride-sharing. 

\subsection{Arc formulation}\label{sec:compact}

We first introduce a direct formulation of the MMCRP
based on the auxiliary graph $G=(V,E)$ presented in Section~\ref{sec:probform}.
For every node $v \in V$ we have the set of outgoing arcs $E_v^+$ and ingoing arcs $E_v^-$.
Let $V'$ be the set of intermediate nodes $V' = V \setminus \{ (d,\sigma), (d,\tau) \}_{d\in D}$.
The set $E^{q}$ denotes all arcs $e$ that cover task $q$, including co-riding visits. $Q$ denotes the set of all tasks.
Finally let $\gamma_e$ denote the savings of arc $e$.
The binary decision variable $x_e$ takes on value 1 if arc $e$ is selected in the solution and 0 otherwise. 

\begin{eqnarray}
\mbox{max}	\label{eq:OF}
			& \quad \sum\limits_{e \in E} \gamma_e  x_e \\
\mbox{s.t} 
 			\label{eq:flowcons}
			&\sum\limits_{e \in E_v^-} x_e = \sum\limits_{e \in E_v^+} x_e  &\forall v \in V'  \\
			\label{eq:depotstart}
			&\sum\limits_{e \in E_{(d,\sigma)}^+} x_e = W_d  &\forall d \in D \\
			\label{eq:depotend}
			&\sum\limits_{e \in E_{(d,\tau)}^-} x_e = \overline{W}_d  &\forall d \in D  \\
			\label{eq:tasks}
			&\sum\limits_{e \in E^{q}} x_e \leq 1  &\forall q \in Q \\
			\label{eq:time}
			& x_e \in \{0,1\}	&\forall e \in E 
\end{eqnarray}
\noindent
The objective (\ref{eq:OF}) maximizes total savings over all arcs. 
Constraints (\ref{eq:flowcons}) 
ensure flow conservation at intermediate nodes $v \in V'$.
Constraints~(\ref{eq:depotstart}) and (\ref{eq:depotend}) ensure that there is a correct number of vehicles $W_d, \overline{W}_d$ at start and end of the time horizon for each depot $d \in D$.
Constraints~(\ref{eq:tasks}) make sure that each task $q \in Q$ is covered at most once. If a given task is not covered, the assigned user will reach the task using the cheapest other MOT.

The model has $O(E)$ variables, and $O(V+D)$ constraints. Hence it is polynomial in the size of the graph. However, the graph $G=(V,E)$ may be large (exponential in the original input size) due to the number of possible co-rides.

Despite the compact arc formulation, but due to the size of the graph, we will see in the computational experiments that only relatively small problems can be solved using this model. We will therefore introduce a stronger but larger formulation based on a path formulation.

\subsection{Path formulation}\label{sec:path}

In order to introduce a path formulation of the MMCRP, we assume that all possible routes $\rho$ of {all} vehicles are enumerated in the set $\cal R$. Each route $\rho$ must start in node $(d,\sigma)$ and finish in a node 
$(d,\tau)$, traversing arcs $E$ in the auxiliary graph $G=(V,E)$. The start and end 
depots $d$ may be different.

Let $\gamma_{\rho}$ be the savings of route $\rho$ calculated as the saving by using a car compared to the cheapest other MOT for all arcs on the respective route.
Furthermore, let the binary matrix $\myG_{\rho q}$ be 1 if route $\rho$ will service task $q$. 
Finally, let $\myH_{\rho d}=1$ if route $\rho$ starts in depot $d$, and $\myQ_{\rho d}=1$ if route $\rho$ ends in depot $d$ and 0 otherwise. 
The values $W_d$ and $\overline{W}_d$ state the number of vehicles that are available at depot $d \in D$
at the beginning and end of the planning horizon.
The binary decision variable $x_{\rho}$ takes on value 1 if route $\rho$ is chosen, and 0 otherwise.

We can now formulate the MMCRP as follows:	\vspace*{-1mm}
\begin{eqnarray}
  \mbox{max} & \sum\limits_{\rho \in \cal R} \gamma_{\rho} x_{\rho} 
               \label{objective} \\
  \mbox{s.t} & \sum\limits_{\rho \in \cal R} \myG_{\rho q} x_{\rho} \leq 1 &
               q \in Q \label{atmostone}\\ 
             & \sum\limits_{\rho \in \cal R} \myH_{\rho d}x_{\rho} = W_d &
               d \in D \label{depotout} \\
             & \sum\limits_{\rho \in \cal R} \myQ_{\rho d}x_{\rho} = \overline{W}_d &
               d \in D \label{depotin} \\
             & x_{\rho} \in \{0,1\} &
               \rho \in \cal R \label{domain}
\end{eqnarray}
The objective function (\ref{objective}) maximizes the sum of the savings of the selected 
routes. 
Constraints~(\ref{atmostone}) make sure that each task $q\in Q$ is covered at most once. 
Constraints~(\ref{depotout}) make sure that exactly $W_d$
vehicles leave depot $d$ at the start of the planning horizon, and 
constraints~(\ref{depotin}) make sure that exactly $\overline{W}_d$
vehicles return to depot $d$ at the end of the planning horizon.
Since not all vehicles have to be used, we add the necessary dummy routes to the set $\cal R$.
Finally, constraints~(\ref{domain}) define the decision variables $x_{\rho}$ 
to be binary.

\subsection{Delayed column generation}\label{sec:delayed}

Since the number of routes $\cal R$ in model 
(\ref{objective})-(\ref{domain}) may be very large we solve its
LP-relaxation through column generation. 

The restricted master problem considers a subset of routes $\cal R' \subseteq R$ of all 
possible routes. In every iteration a pricing problem is solved to find a new
route with positive reduced savings. The process is repeated until no more routes
with positive reduced savings can be found. When the process terminates, we
have solved the LP-relaxation of (\ref{objective})-(\ref{domain}) and hence
have an upper bound on the solution to the MMCRP.

The pricing problem is searching for a route through the auxiliary graph $G=(V,E)$ maximizing the reduced savings.
The problem becomes a kind of (time constrained) shortest-path problem in $G$ where we aim at finding the paths with the largest savings.
Since $G$ is a time-space network, the time constraints are implicitly 
handled by graph construction. Moreover, we note that $G$ is a DAG, and 
hence no cycles can occur.

Let $\dualalpha_{q}$ be the dual variable corresponding to task covering constraint~(\ref{atmostone}), $\dualbeta_{d}$ be the dual variable corresponding to depot start-inventory constraint~(\ref{depotout}), and $\dualgamma_{d}$ be the dual variable corresponding to depot end-inventory constraint~(\ref{depotin}). The reduced savings of a route $\rho$ starting at depot $d$ and ending at depot $d$ can be calculated as follows:
\begin{eqnarray}
  \sum_{q \in \rho}  (\gamma_{q} - \dualalpha_{q}) - \dualbeta_{d} - \dualgamma_{d} 
\end{eqnarray}
The function sums all savings $\gamma_q$ of tasks $q$ covered by route $\rho$ subtracted by the dual variables $\dualalpha_{q}$. Finally the dual variables $\dualbeta_{d}, \dualgamma_{d}$
corresponding to the depot inventory constraints are subtracted.

\subsection{Pricing problem}\label{sec:pricing}

The pricing problem generates new promising routes by finding a path with the maximum reduced savings in the time-space network $G=(V,E)$. We find the path 
using a label setting algorithm. 

The pricing problem is solved for each combination of start depot $d \in D$ and end depot $d \in D$. 
Promising routes with positive reduced savings are added to the master problem until no more routes with positive reduced savings can be found. 

\subsubsection{Dynamic programming}\label{sec:dynprog}

We solve the pricing problem for every pair of start and end
depot, using auxiliary graph $G=(V,E)$. We use a label setting
algorithm adapted to a DAG.
For every node $v \in V$, we have an associated
value $f_v$ denoting the path with the so far largest savings to $v$.
Initially $f_v = -\infty$ for all nodes $v \in V$ except the source
node, and gradually the value of $f_v$ is increased as paths with higher savings
are encountered.

{A dynamic program is solved for} each pair of depots. However, for each start depot, the dynamic programming algorithm will actually solve the problem for all destination depots. So, we only need to call the dynamic programming algorithm $|D|$ times, resulting
in the overall time complexity $O(|D| \times |E|)$ for solving all pricing problems.

Since we have one value $f_v$ for every node, the space complexity of the
dynamic programming algorithm is $O(|V|)$. This is clearly overshadowed
by the size of the graph.

Notice that when the dynamic programming algorithm terminates we may have
several distinct solutions ending at depot $d$, for different arrival times $h$.
The algorithm may easily be modified to return all these distinct solutions.

\subsubsection{Stopping criterion and columns added}\label{sec:multiple}

In the basic pricing algorithm we run the pricing for each combination of depots such that the column with the most positive reduced savings is added. We denote this strategy as \best. Alternatively, we also evaluate the following strategies: \first, \firstdep, and \multiple. In \first\ we stop as soon as a column with positive reduced savings has been found and add this column to the master problem. Again, in each iteration only one column is added. Next, we extend \first\ for every depot combination, and we iterate until the first column with positive reduced savings is found for each combination and terminate thereafter. This means that, when considering two depots and combining each of them, we have at most 4 columns added in each iteration. This is denoted as \firstdep. Lastly, in \multiple\ we include all columns with positive reduced savings. Note that we also tried to restrict the number of columns added. However, we did not see a remarkable difference to the non-restricted case.

\subsubsection{Heuristic pricing algorithm}\label{sec:heur}

Although the pricing problem is solvable in polynomial time in the size of the graph $G = (V,E)$, the number of arcs $E$ may be very large, and we will see in the computational experiments that the pricing problem takes up most of the solution time. We therefore introduce a number of heuristic pricing algorithms, namely \statespace, \heurprun\ and \heuredges. When employing one of the heuristic pricing strategies, we search for columns with positive reduced savings and afterwards finish with one of the exact pricing schemes.

\statespace: In this method, we reduce the auxiliary graph by merging nodes with similar time in the time-space graph. The time-horizon
is discretized in intervals of 10 minutes. If two nodes, corresponding to the same depot, end up in the same time interval, they are merged.
After the merging, there may be multiple arcs between each pair of nodes, so we select the arc with highest savings, and ignore the rest. 

\heurprun:
In this method, we use an aggressive reduction of the graph, by only keeping the savings and not the time. Hence, for every set of ride arcs $E_p = \{ \left( (a_p^{\pi}, g_p^{\pi}) , (b_p^{\pi}, h_p^{\pi}) \right) \}_{{\pi} \in R}$ we merge start and end times, $g_p^{\pi}$ and $h_p^{\pi}$, into one common artificial time $g'_p$ and $h'_p$ for each user's start and end depot $a_p$ and $b_p$ such that, to start with, we only keep track of the ride with largest savings ${{\pi} \in R}$ for each user $p$. 

\heuredges:
In the original algorithm we construct an auxiliary graph $G=(V,E)$ in which we may have arcs $e \in E$ with both positive and
negative savings $\gamma_e$. {A}rcs having a negative saving will only be used if they can be combined with arcs having
a positive saving. The heuristic pricing algorithm removes all arcs $e$ with negative savings $\gamma_e < 0$ before 
running the label setting algorithm. This reduces the size of the auxiliary graph.

\section{Computational study} \label{sec:comp}

The algorithms are implemented in C/C++ and for the solution of the master problem CPLEX 12.6.2 together with Concert Technology 2.9 is used. Tests are carried out using one core of an Intel Xeon 2643 machine with 3.3 GHz and 16 GB RAM running Linux CentOS 6.5. The algorithms are tested on a number of generated instances of increasing size and complexity. Various pricing schemes are compared, and the efficiency of all parts of the code is evaluated. {Most of the reported computation times include the generation of the auxiliary graph. The exception is the comparison of the arc formulation to the column generation approach, where only the time to solve the models is stated.}

To start the column generation we provide an initial set of dummy columns by inserting route variables $x_{\rho}$ such that the master problem is feasible without considering any valid route construction. These dummy variables are leaving and entering {a} depot (constraints \eqref{depotout}-\eqref{depotin}), but do not cover any tasks. We first solve the linear relaxation of the master problem and thereafter we solve the restricted master problem to integer optimality, using only routes generated in $\cal R'$, containing a subset of all routes $R$ with positive reduced savings. In this way we get an upper bound from the column generation based approach, and a lower bound from solving the IP model. Although we cannot guarantee an optimal solution to the original MMCRP in this way, the results will show that in most cases the gap is very small, and the solution quality is more than sufficient for practical applications. 

In the following, we first introduce the test instances in Section~\ref{sec:data}. In Section~\ref{sec:comparisonpricing} we compare the pricing schemes introduced in Section~\ref{sec:pricing} and afterwards conduct algorithmic tests in Section~\ref{sec:algorithmictests}. Finally, we discuss socio-economic aspects in Section~\ref{sec:socio}. For a better understanding of the problem we provide a sample solution in the Appendix~A.

We show in our results that the column generation approach is an efficient choice to solve the MMCRP. We can solve the biggest instances with 300 users and 40 vehicles within
less than one hour on average. Pricing strategy \multiple\ turns out to be the most efficient of the exact methods. The heuristic approaches (\statespace, \heurprun, \heuredges) do not come with a significant improvement in solution time or quality.
Finally, we compare our results to those of the direct formulation presented in Section~\ref{sec:compact}.

\subsection{Test instances}\label{sec:data}

We generate realistic benchmark instances based on available demographic, spatial and economic data of the city of Vienna, Austria. Five different MOTs are considered: car, walk, bike, public transport, and taxi. Walk, bike, public transport, and taxi are assumed to have an unrestricted capacity $m_k$ = $\infty$, while there is a limited number $m_{car} < \infty$ of shared cars.
For each mode of transport $k \in K$ we define a set of properties, described in the following. Information of the car is based on available data (PKW-Mittel Diesel in \citet{QuoVadis2010}). 
Distance $d_{ij}^k$, time $w_{ij}^k$ and cost $c_{ij}^k$ are calculated between all nodes $i$ and $j$ for all modes of transport $k \in K$. Average travel speed per transport mode are as follows (in km/h): car = 30, walk = 5, bike = 16, public transport = 20, taxi = 30. 

Emissions $\epsilon^k_{ij}$  are translated into costs and, together with distance cost $c^k_{ij}$ and cost of time $w'^{k}_{ij}$, included into the overall cost calculations. Costs per emitted ton of CO$_2$ is 5\euro{} and average gross salary in Austria including additional costs for the employer is 19.42\euro{}/hour. Variable cost per distance $c^k_{ij}$ of the car is taken from the available car information in \citet{QuoVadis2010} and is 0.188\euro{}/km. For taxi we take on a value of 1.2\euro{}/km. As we only consider distance cost that are variable and no fixed charges, we assume for all other MOTs costs $c^k_{ij}$ of 0. 
Additional time $\xi^k$ is added to denote extra time needed for a certain MOT $k$, such as additional 10 minutes for cars to account for walking distances from/to the parking lot. This we specify as follows (in seconds): $\xi^{car}$ = 600, $\xi^{walk}$ = 0, $\xi^{bike}$ = 120, $\xi^{public}$ = 300, $\xi^{taxi}$ = 300. Distances are based on aerial distances and multiplied by a constant sloping factor $\zeta^k$ for each mode $k$ in order to account for shortcuts/detours usually associated with certain modes of transport. These we define as $\zeta^{car}$ = 1.3, $\zeta^{walk}$ = 1.1, $\zeta^{bike}$ = 1.3, $\zeta^{public}$ = 1.5, and $\zeta^{taxi}$ = 1.3.

Each generated instance represents a distinct company consisting of one or more depots $d \in D$ and users, i.e.{,} employees, $p \in P$. The locations of the depots are based on statistical data of office locations in Vienna. The set of possible locations is based on geometric centers of all 250 registration districts of Vienna.

Companies are defined by a fixed number of users $u$ and depots $|D|$. The number of customer visits, i.e., meetings, and their time and location, are then randomly generated based on historic statistical data.

To each user $p$ we associate a subset of MOTs $K^p \subseteq K$. We assume penalties for constraint violations such as choosing a mode of transport that is not in the user's choice or for too late arrival. The penalty cost per violation is determined to be 10,000 and directly included into the cost function.

For our calculations, we depict one day only. Each user $p$ has an assigned set of tasks $Q^p$ distributed over the day. For the smallest instances on average 95 nodes (comprising meetings/tasks $q$ and start and end depots $a,b$) and 33 tasks are generated. For the largest instances we have on average 463 tasks. Further information is given in Section~5.4 in Table~\ref{tab:numsimedge}. This leads to about 4-5 assigned nodes per user (including their start and end depots of the trips).
The ordered list of tasks $Q^p$ for a working day per user $p$ is generated with the following attributes for each task $q$: latest arrival, earliest departure, service duration, all given in minutes. We already account for ride-sharing in the instance generation where we enforce the proximity of various tasks of different users. 
First, a predefined sequence of tasks is generated per user $p$ which is then partitioned into separate sets of tasks with newly assigned artificial user $p' \in P$ if a user returns to the depot more than once during a day. If a user $p$ has more than one simple trip, buffer time at the depot is set to 60 minutes in order to account for, e.g., changing of cars or additional time needed if the previous route was not covered by car. We assume that the maximum distance between two nodes is one hour.

Instances are named as \texttt{E\_}$u$\texttt{\_}$I$,
where $u$ is the number of users, and the instance number $I$ is between 0 and 9. For example, the first instance in 
the set of instances with 20 users ($u = 20$) is denoted \texttt{E\_20\_0}. For each combination of $u$ and $m$ 
we solve a set of 10 instances (\texttt{E\_u\_0} to \texttt{E\_u\_9}) and report the average values.

Instance sets and the source code of the instance generators are made publicly available at \url{https://github.com/dts-ait/seamless}.

\subsection{Comparison of the different pricing schemes}\label{sec:comparisonpricing}

In this part, we compare different pricing schemes and study heuristic pricing algorithms. We compare four different variants of how and when to add columns to the master problem (described in Section~\ref{sec:multiple}) and three heuristic approaches, as described in Section~\ref{sec:heur}. We provide insights into different parts of the algorithm and finally choose the variants having the best trade-off between solution time and solution quality.
We compare results based on an increasing number of users $u=20,50,100,150,200,250,300$ and vehicles $m=2,4,10,20,40$. 
In our experiments the number of depots is two except for Table~\ref{tab:diffdepot} where we study instances with more depots. The vehicles are equally split over all depots. 

To start with, we study the solution time and solution quality of the different exact pricing schemes described in 
Section~\ref{sec:multiple}. Notice that all pricing schemes return the same LP-bound, but the IP-solution may
be different because a different set of columns may be generated.

In Table~\ref{tab:cpuexact} we report computational times in seconds and the average gap in percentages between the integer and LP solution for the respective set of instances. The gap between the two solutions is calculated as: $(\mbox{Savings LP} - \mbox{Savings IP}) / \mbox{Savings IP}$.  We split the table into combinations of pricing scheme (\best, \first, \firstdep, \multiple), number of vehicles ($m = 2,4,10,20,40$) and users ($u=150,200,250$). Note that for \multiple\ we add a row for $u=300$ as this is the only exact scheme that was able {to} solve all instances within the stated time limit of two hours. As it may be seen, we are able to find near optimal solutions with a gap close to $0$. For the case with $m = 2$, which means one vehicle for each depot, we can close the gap for all instances given in the table. The gap increases slightly when more vehicles are added, however only up to a certain point. For instances with more cars ($m = 10,20,40$) we cannot see a significant difference in the gap anymore. All approaches return solutions of approximately equally good quality. However, strategy \multiple\ gives slightly better gaps than the other schemes. In schemes \best, \first, and \firstdep\ only a very restricted subset of columns is added to the problem in each iteration. However, they might not be the most beneficial for the integer solution. With \multiple\ we add all found columns with positive reduced savings which helps us to identify an integer solution. Given this situation and also for practical reasons, we decided to refrain from implementing a full fledged branch-and-price algorithm.
In terms of computational time, for instances with a small number of cars ($m = 2,4$) and up to 150 users ($u = 150$) it does not make a difference which scheme is used. By increasing the number of users and keeping a small fleet we can gradually see a difference. Pricing scheme \best\ performs worst and pricing scheme \multiple\ is, by far, the most efficient in terms of computation time. For the largest comparable instance which can be solved by all schemes ($m = 40$ and $u = 250$), computation times differ by a factor of 7: the average run time of pricing scheme \multiple\ amounts to 843 seconds and to 6,145 seconds with pricing scheme \best. The biggest instances ($u = 300, m = 40$) can be solved within less than one hour on average.
We can observe in our computational results that the time spent on solving the master problem is usually very small, below three minutes on average, with the exception of instance class $u = 100$ and $m = 10$, where we obtain an average value of 1,296 seconds, using pricing scheme \first. Most of the computation time is spent on solving the pricing problem. Observing this, we try to decrease computation times by studying three different heuristics to accelerate pricing, namely \heuredges, \heurprun, \statespace\ (see Section~\ref{sec:heur}). 

\begin{table}[h!]
  \centering
  \renewcommand{\baselinestretch}{1.0}
 \caption{Average computation time in seconds and average gap {in percentages} (\%) between the LP solution and the integer solution for the exact pricing schemes for an increasing number of users ($u$) and vehicles ($m$). The gap is calculated as: $(\mbox{Savings LP} - \mbox{Savings IP}) / \mbox{Savings IP}$.}
   \renewcommand\baselinestretch{1}\selectfont
   \setlength{\tabcolsep}{4pt}
    \begin{tabular}{crrrrrrrrrrrrrrrrr}
    \toprule
       &  &   & \multicolumn{2}{c}{\textit{m} = 2} &   & \multicolumn{2}{c}{\textit{m} = 4} &   & \multicolumn{2}{c}{\textit{m} = 10} &   & \multicolumn{2}{c}{\textit{m} = 20} &   & \multicolumn{2}{c}{\textit{m} = 40} \\
      & \multicolumn{2}{l}{\textit{u}} &  \multicolumn{1}{c}{time} & \multicolumn{1}{c}{\%} &   & \multicolumn{1}{c}{time} & \multicolumn{1}{c}{\%} &   & \multicolumn{1}{c}{time} & \multicolumn{1}{c}{\%} &   & \multicolumn{1}{c}{time} & \multicolumn{1}{c}{\%} &   & \multicolumn{1}{c}{time} & \multicolumn{1}{c}{\%} \\
\cmidrule{2-17}    \multirow{3}[1]{*}{\best} & 150 &   & 4.7 & 0.00 &   & 17.3 & 0.46 &   & 97.1 & 1.08 &   & 421.9 & 0.25 &   & 845.6 & 0.14 \\
      & 200 &   & 9.1 & 0.00 &   & 37.1 & 0.21 &   & 328.2 & 0.45 &   & 2797.6 & 0.18 &   & 2805.8 & 0.23 \\
      & 250 &   & 18.9 & 0.00 &   & 83.4 &0.07 &   & 1799.7 & 1.05 &  & 2738.6 & 1.07 &  & 6145.4 & 0.18 \\
      &   &   &   &   &   &   &   &   &   &   & & \\
    \multirow{3}[0]{*}{\first} & 150 &   & 4.1 & 0.00 &   & 12.1 & 0.46 &   & 58.8 & 0.83 &   & 254.4 & 0.23 &   & 582.6 & 0.08 \\
      & 200 &   & 7.6 & 0.00 &   & 28.9 & 0.21 &   & 355.3 & 0.43 &   & 1396.7 & 0.22 &   & 2058.1 & 0.15 \\
      & 250 &   & 16.5 & 0.00 &   & 61.7 & 0.07 &   & 1321.7 & 0.79 &   & 3512.2 & 1.11 &   & 4506.4 & 0.15 \\
      &   &   &   &   &   &   &   &   &   &   & & \\
    \multirow{3}[0]{*}{\firstdep} & 150 &   & 4.4 & 0.00 &   & 14.6 & 0.47 &   & 86.7 & 0.98 &   & 242.2 & 0.26 &   & 518.5 & 0.11 \\
      & 200 &   & 7.7 & 0.00 &   & 28.4 & 0.21 &   & 277.8 & 0.49 &   & 1307.4 & 0.22 &   & 1690.5 & 0.18 \\
      & 250 &   & 16.9 & 0.00 &  & 64.3 & 0.07 &   & 625.6 & 0.98 &   & 1677.9 & 1.18 &   & 3303.6 & 0.16 \\
      &   &   &   &   &   &   &   &   &   &   &  & \\
    \multirow{4}[1]{*}{\multiple} & 150 &   & 4.0 & 0.00 &   & 9.3 & 0.46 &   & 28.3 & 0.63 &   & 72.2 & 0.24 &   & 114.6 & 0.04 \\
      & 200 &   & 8.3 & 0.00 &   & 20.4 & 0.21 &   & 108.3 & 0.44 &   & 270.7 & 0.23 &   & 362.6 & 0.10 \\
      & 250 &   & 17.7 & 0.00 &   & 42.3 & 0.07 &   & 222.1 & 0.57 &   & 493.7 & 0.75 &   & 842.7 & 0.12 \\
     & 300 & & 50.1 & 0.00  & & 156.4 & 0.70 & & 631.5 & 0.59 & & 1734.4 & 0.11 & & 2994.6 & 0.16 \\
    \bottomrule
    \end{tabular}%
  \label{tab:cpuexact}%
\end{table}%

All heuristic approaches use pricing scheme \multiple\ as it turned out to be the most efficient of the introduced exact pricing algorithms. The heuristic pricing schemes are employed as follows: We use the heuristic as long as columns with positive reduced savings can be found. Thereafter, we continue with the chosen exact pricing procedure.

Table~\ref{tab:heuristics} gives the average gap between the upper bound and solving the original IP on the same set of columns, and the average computation times in seconds for the heuristic pricing schemes \heuredges, \heurprun, \statespace, and scheme \multiple\ for the larger instance classes defined by $u = 150,200,250,300$ and $m = 2,4,10,20,40$. We see for all schemes a gap below 1\%, and similar computational times. Thus, none of the presented schemes significantly stands out in terms of running times or solution quality. 

\begin{table}[h!]
  \centering
  \renewcommand{\baselinestretch}{1.0}
  \caption{Average computation time in seconds and average gap {in percentages} (\%) between the LP solution and the integer solution for the heuristic pricing schemes and scheme \multiple\ for $u=150,200,250,300$ and an increasing number {of vehicles} $m$. The gap is calculated as $(\mbox{Savings LP} - \mbox{Savings IP}) / \mbox{Savings IP}$ and reported for increasing number of vehicles ($m$).}
  \renewcommand\baselinestretch{1}\selectfont
    \begin{tabular}{lrrrrrrrrrrrrrr}
    \toprule
          & \multicolumn{2}{c}{$m$ = 2} &       & \multicolumn{2}{c}{$m$ = 4} &       & \multicolumn{2}{c}{$m$ =10} &       & \multicolumn{2}{c}{$m$ = 20} &       & \multicolumn{2}{c}{$m$ = 40} \\
          & \multicolumn{1}{c}{time} & \multicolumn{1}{c}{\%} &       & \multicolumn{1}{c}{time} & \multicolumn{1}{c}{\%} &       & \multicolumn{1}{c}{time} & \multicolumn{1}{c}{\%} &       & \multicolumn{1}{c}{time} & \multicolumn{1}{c}{\%} &       & \multicolumn{1}{c}{time} & \multicolumn{1}{c}{\%} \\
\cmidrule{2-15}    \multiple & 20.0  & \textbf{0.00}  &       & 57.1  & 0.59  &       & 247.5 & \textbf{0.45}  &       & 642.8 & \textbf{0.21}  &       & 1078.6 & 0.10 \\
    \heuredges & \textbf{18.4}  & \textbf{0.00}  &       & 48.5  & 0.21  &       & \textbf{218.7} & 0.53  &       & \textbf{571.1} & 0.44  &       & 1075.2 & 0.10 \\
    \heurprun & 22.2  & \textbf{0.00}  &       & 52.3  & 0.20  &       & 229.4 & 0.57  &       & 616.6 & 0.50  &       & \textbf{1036.7} & \textbf{0.09} \\
    \statespace & 19.1  & \textbf{0.00} &       & \textbf{47.6}  & 0.23  &       & 224.9 & 0.57  &       & 571.9 & 0.43  &       & 1060.7 & 0.11 \\
    \bottomrule
    \end{tabular}%
  \label{tab:heuristics}%
\end{table}%

In Table~\ref{tab:colsgen}, we present the total number of columns generated when running the respective pricing algorithm over all instances with $m = 2,4,10,20,40$. We observe that, as expected, by only adding one column in each iteration (\best, \first) and one for each depot combination (\firstdep) we generate fewer columns than with scheme \multiple, which adds all columns with positive reduced savings. 

\begin{table}[h!]
  \centering
  \renewcommand{\baselinestretch}{1.0}
  \caption{Total number of columns generated by the different pricing schemes over all instance classes with a given number of vehicles ($m$).}
    \renewcommand\baselinestretch{1}\selectfont
    \begin{tabular}{lrrrrrr}
    \toprule
      &   & \multicolumn{1}{l}{\textit{m }= 2} & \multicolumn{1}{l}{\textit{m} = 4} & \multicolumn{1}{l}{\textit{m }= 10} & \multicolumn{1}{l}{\textit{m} = 20} & \multicolumn{1}{l}{\textit{m} = 40} \\
\cmidrule{3-7}    \best &   &             28    &          149    &          855    &       2,771    &       4,993    \\
    \first &   &             30    &          163    &    13,903    &       3,883    &       6,606    \\
    \firstdep &   &             29    &          154    &          857    &       2,585    &       4,710    \\
    \multiple &   &          263    &          645    &       2,751    &       6,797    &    12,865    \\
    \bottomrule
    \end{tabular}%
  \label{tab:colsgen}%
\end{table}%

The solutions {gained} when solving the arc formulation and the integer solutions obtained by the column generation algorithm are compared in Table~\ref{tab:edge}. We run the arc formulation on a set of small instances ($u = 20, 50, 100$ and $m = 4$) already showing the benefits of the decomposition algorithm. In the first row ({computation time} (s)) the average {computation} times {to solve} the arc formulation (AF), and the column generation algorithm (CG) are given. The stated {numbers} encompass {only the computation times in seconds to solve the respective formulation}, thus the enumeration of the trips is excluded. We observe that for the smallest instances the arc formulation is faster. However, for the bigger instances ($u=50,100$), we need a multiple of the time of the column generation algorithm.
The gap (gap (\%)) between the solution of the arc formulation and the respective integer solution is very low, less than 0.11\% for all instances on average. With the column generation approach we are able to solve 9 out of 10 instances of the respective instance group to optimality (\# opt.)

As we assumed and also {show} in Table~\ref{tab:numsimedge}, the number of arcs generated, and thus the underlying graph, becomes very large and therefore this formulation cannot be used efficiently in a direct formulation. Note that we also tried to run the arc formulation by increasing the relative MIP gap (e.g.{,} to 1\%). This, however, did not lead to any improvements as solving the root relaxation takes most of the computational time.

\begin{table}[h!]
  \centering
  \renewcommand{\baselinestretch}{1.0}
  \caption{Average time in seconds for solving the arc formulation and the column generation based algorithm, average gap {in percentages} between the solution of the arc formulation (AF) and integer solution of the column generation based algorithm (CG), and number of instances solved to optimality with the column generation based algorithm for instances with {an increasing number of users} $u=20,50,100$ and {number of vehicles} $m=4$.}
  \renewcommand\baselinestretch{1}\selectfont
    \begin{tabular}{lrrrrrr}
    \toprule
    \textit{u=} & \multicolumn{2}{c}{20} & \multicolumn{2}{c}{50} & \multicolumn{2}{c}{100} \\
    \midrule
          & \multicolumn{1}{c}{AF} & \multicolumn{1}{c}{CG} & \multicolumn{1}{c}{AF} & \multicolumn{1}{c}{CG} & \multicolumn{1}{c}{AF} & \multicolumn{1}{c}{CG} \\
\cmidrule{2-7}    {computation time} (s) &                   0.1  &                   0.6  &                23.1  &                   0.8  &          1,615.9  &                   1.2  \\
     gap (\%) & 0.00 & {0.06} & 0.00 & {0.11} & 0.00 & {0.03} \\
    \#opt. &    10/10   & 9/10  &   10/10  & 9/10  &   10/10    & 9/10 \\
    \bottomrule
    \end{tabular}%
  \label{tab:edge}%
\end{table}%

\subsection{Algorithmic tests}\label{sec:algorithmictests}

In the following we use configuration \statespace\ to study how the column generation evolves, the impact of early termination of the column generation and different numbers of depots.

In terms of convergence of the algorithm, we observe the common picture of a steep increase in the objective value during the first iterations and then a long tail until optimality of the LP relaxation has been proven. This means that we are able to find good solutions close to the optimal objective function value in a relatively few iterations. However, the column generation then needs quite a number of iterations in order to find the optimal value. This effect can be exploited for practical applications: the column generation process can be terminated early without loosing much in terms of solution quality.
Figures~4(a), (b), (c) and (d) in the Appendix {B} plot the convergence of the column generation algorithm. The number of iterations is shown on the x-axis and the objective function value on the y-axis. We report results for $u=150$ users (a), $u=200$ user (b), $u=250$ users (c), and $u=300$ users (d). Each curve represents the outcome of one instance. 

In Table~\ref{tab:stopiter} the impact of early termination after about one third of the iterations on the quality of the obtained objective value is shown. As the number of iterations needed to find the optimal solution does not vary much between instances of different size, we assume a common termination criterion for all. Observing that usually about 140-160 iterations are needed to terminate the algorithm, we study early termination after 50 iterations and solve the integer problem on the columns generated so far. Table~\ref{tab:stopiter} gives the gap in percentages between the integer and LP solution for each instance with $u=300$. The row ''time {(s)}'' shows the average run time in seconds. 
We observe that we are still able to find good solutions after only one third of the iterations. The computed gap between the original upper bound and solving the IP on the restricted set of columns, using early termination, is at most 6.6\% and 2.5\% on average, which is good enough for practical applications. Since the algorithm is stopped after about one third of the previously necessary iterations, run times are reduced accordingly.

\begin{table}[h!]
\scriptsize
\renewcommand{\baselinestretch}{1.0}
  \centering
  \caption{Impact of early termination of the column generation algorithm after 50 iterations for $u=300$ and $m=40$. Comparison between original upper bound and obtained integer solution after early termination. Gap {in percentages} and time in seconds are given for the case of early termination (terminate) and the original values obtained from \statespace.}
  \setlength{\tabcolsep}{0.3em}
    \begin{tabular}{clrrrrrrrrrrr}
    \toprule
          &       & \multicolumn{1}{l}{E\_300\_0} & \multicolumn{1}{l}{E\_300\_1} & \multicolumn{1}{l}{E\_300\_2} & \multicolumn{1}{l}{E\_300\_3} & \multicolumn{1}{l}{E\_300\_4} & \multicolumn{1}{l}{E\_300\_5} & \multicolumn{1}{l}{E\_300\_6} & \multicolumn{1}{l}{E\_300\_7} & \multicolumn{1}{l}{E\_300\_8} & \multicolumn{1}{l}{E\_300\_9} & \multicolumn{1}{c}{average} \\
\cmidrule{3-13}  \multirow{2}[2]{*}{gap (\%) }  & terminate &              6.45  &              2.56  &             2.29  &              0.19  &              2.70  &             2.82  &              0.16  &             1.80  &              4.52  &              2.55  &              2.60  \\
          & original &                   0.00    &              0.76  &             0,78  &                   0.00    &              0.72  &                  0.00    &                   0.00    &             0.61  &                   0.00    &              0.06  &              0.16  \\
    \multirow{2}[2]{*}{time (s)} & terminate &            637.9  &            506.3  &           101.1  &            958.2  &        1,193.0  &           119.7  &            905.7  &             69.8  &        1,325.3  &        1,101.3  &            691.8  \\
          & original &        5,533.2  &        3,198.1  &           424.8  &        1,043.4  &        5,430.3  &           402.5  &        1,051.5  &           328.2  &        7,340.5  &        5,194.0  &        2,994.6  \\
    \bottomrule
    \end{tabular}%
  \label{tab:stopiter}%
\end{table}%

In Table~\ref{tab:diffdepot} we show the impact of increasing the number of depots on the run time of the algorithm. In all previous experiments, two depots were used. We now use 1, 3, and 4 depots. Within the project, the case of 1 depot is chosen as currently a company usually operates one main office (or at maximum two) with shared cars. However, as we are investigating the future sharing economy and different settings of companies we also analyse if our algorithm is capable of dealing with more depots. The number of vehicles is shown as the number of vehicles per depot to allow for a fair comparison. This means that, to obtain the actual total number of vehicles, this number must be multiplied by the number of depots. 
As expected, computation times increase with rising number of depots, however only to a certain factor and not exponentially. As the pricing algorithm is solved for each pair of depots we are increasing the number of subproblems the pricing algorithm is able to handle. However, as only a few trips start and end at different depots, we can provide reasonable solution times even for the case of four depots. In Table~\ref{tab:diffdepot} the respective values are provided.

\begin{table}[h!]
  \centering
  \renewcommand{\baselinestretch}{1.0}
  \caption{Average computation times {in seconds} for 1, 2, 3 and 4 depots for users $u=20,50,100,150,200,250,300$ and vehicle per depot ($m'$).}
  \renewcommand\baselinestretch{1}\selectfont
    \begin{tabular}{lrrrrr}
    \toprule
          & \multicolumn{1}{l}{\textit{m'} = 1} & \multicolumn{1}{l}{\textit{m' =} 2} & \multicolumn{1}{l}{\textit{m' =} 5} & \multicolumn{1}{l}{\textit{m' =} 10} & \multicolumn{1}{l}{\textit{m' =} 20} \\
\cmidrule{2-6}    1 depot & 8.4   & 15.0  & 62.4  & 134.3 & 372.3 \\
    2 depots & 12.3  & 33.7  & 143.4 & 370.0 & 619.2 \\
    3 depots & 15.6  & 54.9  & 208.5 & 586.7 & 652.2 \\
    4 depots & 35.9  & 143.2 & 554.2 & 1105.2 & 1601.6 \\
    \bottomrule
    \end{tabular}%
  \label{tab:diffdepot}%
\end{table}%

\subsection{Socio-economic tests}\label{sec:socio}

In this section, we summarize the results of our socio-economic tests. All results are obtained using the version of our algorithm with the smallest LP-IP gap, which is \multiple. We study savings by ride-sharing, savings by car-sharing and give insights into the instances and strategic decisions regarding the optimal size of the car pool.

Table~\ref{tab:numsimedge} gives the average number of all trips for each instance class and the average number of simple trips. Simple trips are the arcs without any ride-sharing activities going from start node $a$ to end node $b$. As we can see, the number is always slightly higher than the number of users, indicating that a set of users return to the depot during the day and start another sequence of nodes in the observed planning horizon. 
Column ''tasks incl. start/end'' gives the average number of nodes, i.e., tasks and start/end depots of a user's trip, of the instance set. Each user covers about 4-5 nodes, each trip includes on average 3.3 nodes. The column ''tasks'' gives the average number of tasks for each instance set. This number ranges from 33 ($u=20$) to 463 ($u=250$), which gives approximately 1-2 tasks per trip. In the last four columns we give the percentage of tasks and start/end depots covered by a car. Naturally, with a low number of cars ($m=4$) and a high number of users ($u=250$), only a small subset of tasks will be covered by car. The coverage ranges from 4\% to 45\%.

\begin{table}[h!]
  \centering
  \renewcommand{\baselinestretch}{1.0}
  \caption{Number of simple trips, number of all trips, and tasks in {the} auxiliary graph for {an} increasing number of users $u$. The columns on the right give the percentage of nodes including tasks, start and end depots{,} covered by car {for $m=4,8,20,40$}.}
    \renewcommand\baselinestretch{1}\selectfont
    \begin{tabular}{rrrrrrrrrr}
    \toprule
    \multicolumn{1}{l}{$u$} & \multicolumn{1}{l}{simple trips} & \multicolumn{1}{l}{all trips} & \multicolumn{1}{l}{tasks incl. start/end} & \multicolumn{1}{l}{tasks} & \multicolumn{1}{l}{$m=$} & \multicolumn{1}{l}{4} & \multicolumn{1}{l}{8} & \multicolumn{1}{l}{20} & \multicolumn{1}{l}{40} \\
    \midrule
    20  & 31 &        240   & 95 & 33 & & 30\% & 38\% & 41\% & 41\% \\
    50  & 76 &         7,403   & 242 & 90 & & 13\% & 25\% & 42\% & 43\% \\
     100  & 147 &      71,726  & 476 & 183 & & 9\% & 15\% & 35\% & 45\%  \\
    150  & 218 &     551,936  & 714 &  279 & &  6\% & 11\% & 25\% & 43\% \\
    200  & 287 &    1,497,545  & 951  &  381 & & 5\% & 9\% & 20\% & 37\% \\
    250  & 358 &    2,317,145  & 1,179 & 463  & & 4\% & 7\% & 16\% & 30\%  \\
    \bottomrule
    \end{tabular}%
  \label{tab:numsimedge}%
\end{table}%

In Table~\ref{tab:simple+user}, the results obtained from solving the MMCRP are compared to only car-sharing (ratio (1)) and user-dependent car assignment (ratio (2)). The value is given as the savings ratio of MMCRP : car-/ride-sharing.

User dependent car-assignment means that if a user has an assigned car, the selected user will have the car for the whole day and use it for all trips. Moreover, no other user is allowed to use this car and ride-sharing is not possible. 
If only car-sharing is employed users may hand over the cars during the planning horizon so that a car will have different drivers assigned, however, no ride-sharing is allowed. 
All tests are run for an increasing number of vehicles ($m = 2,4,10,20$) and users ($u = 20,50$). Instance specific results are reported and summed up in the row ''average''.

By allowing car- and ride-sharing and thus having a more flexible usage of the car pool rather than a restricted usage during a day, we can have up to 1.7 times higher savings in the planning horizon. This is already shown for small-sized instances. As expected, the more flexible the usage of the cars, the more savings are achieved. Please note that instances E\_20\_4, E\_20\_7, E\_20\_8, and E\_50\_1 are not in the table, meaning that these are not included in the average calculations as they would give a somewhat misleading outcome. This is due to the fact that we assumed penalties for constraint violations such as choosing a mode of transport that is not in the user's choice or for too late arrival. The penalty cost per violation is determined to be 10,000 and directly included into the cost function. Therefore we obtain for some instances very high savings which is mainly due to these penalties included in the objective function. Note that we also excluded values for $m=20$ and $u=20$ as this setting means that more cars than users are provided.

\begin{table}[h!]
\small
\renewcommand{\baselinestretch}{1.0}
  \centering
  \caption{Increase in savings when comparing car-sharing ({without} ride-sharing), user-dependent car-sharing ({i.e., car-sharing without} ride-sharing and a user has a car for the whole day) and MMCRP. Ratio (1) reports the factor of improvement in savings in comparison to car-sharing, ratio (2) reports the enhancement when comparing to the user-dependent car-sharing.}
    \renewcommand\baselinestretch{1}\selectfont
    \begin{tabular}{lrrrrrrrrrrr}
    \toprule
      & \multicolumn{2}{c}{\textit{m} = 2} &   & \multicolumn{2}{c}{\textit{m} = 4} &   & \multicolumn{2}{c}{\textit{m} = 10} &   & \multicolumn{2}{c}{\textit{m} = 20} \\
      & \multicolumn{1}{c}{ratio (1)} & \multicolumn{1}{c}{ratio (2)} &   & \multicolumn{1}{c}{ratio (1)} & \multicolumn{1}{c}{ratio (2)} &   & \multicolumn{1}{c}{ratio (1)} & \multicolumn{1}{c}{ratio (2)} &   & \multicolumn{1}{c}{ratio (1)} & \multicolumn{1}{c}{ratio (2)} \\
      \midrule
    \textbf{$u=20$} & 1.2 & 1.4 &   & 1.2 & 1.4 &   & 1.3 & 1.3 &   & -  &  - \\
    \textbf{$u=50$} & 1.3 & 1.6 &   & 1.5 & 1.6 &   & 1.6 & 1.7 &   & 1.7 & 1.7 \\
    \bottomrule
    \end{tabular}%
    \label{tab:simple+user}%
\end{table}%

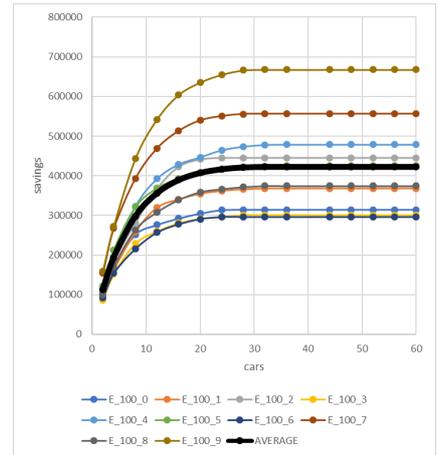
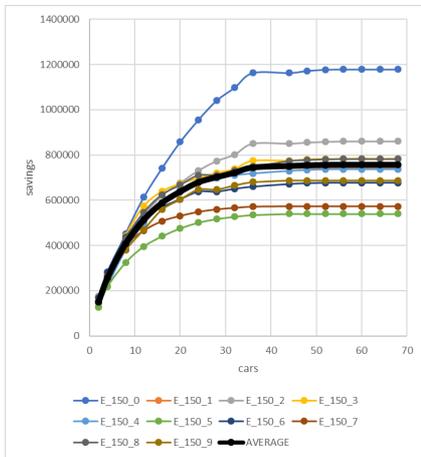
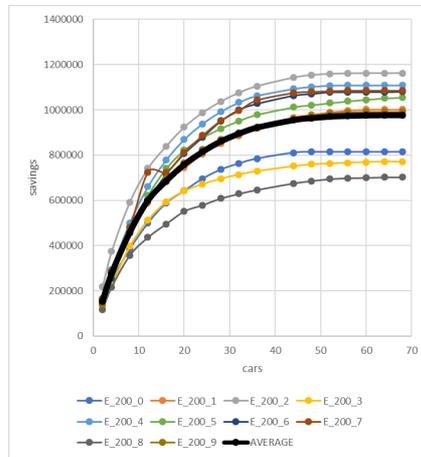
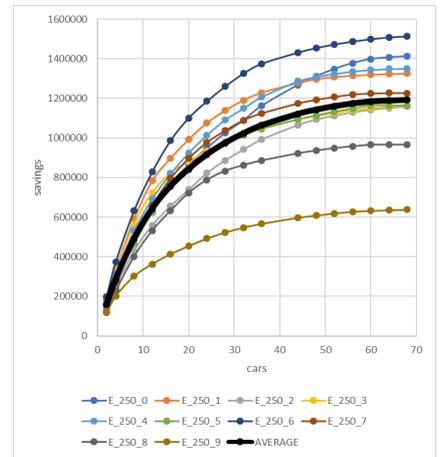
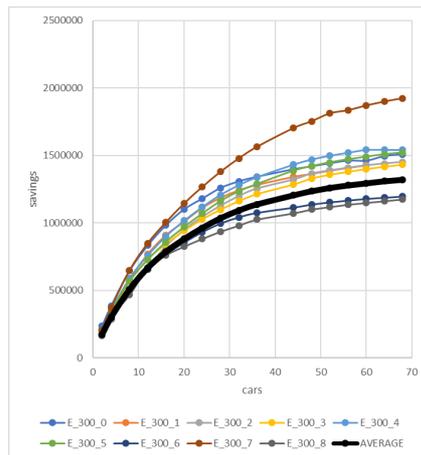
When analyzing fleet size, we see initially big savings when adding more vehicles yet the impact diminishes quite fast. For instances with 20 users fewer than 5 vehicles suffice, for 50 users fewer than 10 are certainly enough and when we consider 100 users the breaking point is somewhere between 20 and 30. For larger instances ($u = 150, 200, 250, 300$), the ideal number of vehicles would be between 20 and 50. Figure~5 in the Appendix {B} provides illustrative insights into the optimal fleet size for an increasing number of users employed ($u = 20,50,100,150,200,250,300$). The x-axis represents number of vehicles, the y-axis the objective function value and each line represents a distinct instance of our experiments, whereas the thicker black line in each subfigure shows the average.

Finally, we analyse the average number of arcs and ride-shares in our results. Table~\ref{tab:rideshares+edges} summarizes the average number of trips per vehicle route and average ride-sharing activities per arc (ride-sharing per ride). This is shown for the cases of 20 and 50 users and increasing number of vehicles ($m = 2,4,10,20$ respectively).
The average number of trips on a vehicle route gives us an idea of the amount of car-sharing activities. With an increasing number of cars the number of trips on a vehicle route is decreasing. The very small numbers, for example 0.9 for $u = 20$ and $m = 10$, are mainly due to unused arcs, meaning that not all of the available cars are used. Moreover, we have on average 1.5-1.7 ride-sharing activities per arc, which is considered to be very high and supports our goal of having a good utilization of the pool of cars.
Note that we again exclude values for $m=20$ and $u=20$ as this setting would mean that more cars than users are provided.

\begin{table}[h!]
  \centering
  \renewcommand{\baselinestretch}{1.0}
  \caption{Average number of trips per car, and average number of ride-shares per trip}
    \renewcommand\baselinestretch{1}\selectfont
    \begin{tabular}{rccccc}
    \toprule
      & \multicolumn{2}{c}{$u = 20$} &   & \multicolumn{2}{c}{$u = 50$} \\
\cmidrule{2-3}\cmidrule{5-6}      & trips per car  & ride-sharings  &   & trips per car & ride-sharings \\
    $m$ & & per trip &   & & per trip \\
\cmidrule{2-3}\cmidrule{5-6}    2 & 2.0 & 1.5 &   & 2.1 & 1.6 \\
    4 & 1.6 & 1.5 &   & 1.9 & 1.5 \\
    10 & 0.9 & 1.2 &   & 1.7 & 1.7 \\
    20 &  - & - &   & 1.1 & 1.7 \\
    \bottomrule
    \end{tabular}%
  \label{tab:rideshares+edges}%
\end{table}%

\section{Conclusion} \label{sec:concl}

Inspired by the concept of sharing economy and future mobility systems, we introduced the multimodal car- and ride-sharing problem (MMCRP) that assigns different modes of transport to ride requests. We aimed at deploying a pool of shared cars as efficiently as possible, join ride requests by offering ride-sharing and by assigning different modes of transport to the remaining requests.

We introduced the novel MMCRP and showed that the problem is NP-hard if the number of depots is at least two. The problem remains NP-hard even if ride-sharing is not allowed. In order to circumvent the complexity of modeling ride-shares and additionally assigning further modes of transport, we constructed an auxiliary graph in which all possible ride-sharing rides are enumerated. Ride requests not covered by a car or ride-share are appointed to take the cheapest other MOT. This made it possible to formulate a compact model for the problem as a kind of vehicle scheduling problem. We extend the vehicle sharing problem by allowing for different start and end depots. Moreover, due to the modeling of ride-sharing into the graph, we may have multiple possibilities to cover a trip and have to make sure that a user is not riding in parallel, {i.e., that a user is not driving in multiple cars at the same time}. Note that the auxiliary graph can be exponential in the original input size. Due to the size of the auxiliary graph, the compact model is also quite complex to solve, hence we proposed a path-based formulation. To solve the path-based formulation we introduced an efficient two-stage decomposition algorithm  and relied on {well-studied} approaches to solve the real-world problem. In the first stage of the decomposition approach, trips were enumerated and afterwards, in the second stage, solved through a column generation approach. The master problem keeps track of the requests and depot balance constraints, while the pricing problem generates improving paths. We showed that the pricing problem can be solved through dynamic programming in polynomial time, measured in the size of the auxiliary graph. The computational results confirmed that large instances can be solved in reasonable time, making it possible to use the algorithm for daily planning of multimodal car- and ride-sharing problems even in a large-scale setup.
The two-level decomposition makes it easy to implement additional constraints on co-riding to make it more attractive: This could be limits on detour or driving time, or co-rider preferences. Also the framework can easily be generalized to handle more than {one} co-rider, which should also be done in future work.

The introduced model targets corporate mobility services. However, it can easily be applied to any specific network with a predefined set of users in a closed community and is therefore of high importance in current and future concepts of shared mobility systems. 
Moreover, the MMCRP can be extended to electric cars, where the algorithm must ensure that sufficient time is available at the depot for recharging the cars. 
Furthermore, the model can be extended to consider several shared modes of transport. This could be electric as well as conventional cars and bikes that are pooled to satisfy the needs of transportation for one or more companies. Bikes might need to be embedded in a rebalancing system, ensuring that the right number of bikes is at the right locations at the beginning and end of the day. Bikes can thus be implemented in a unified model with other shared MOTs considering specific limitations as they cannot be available for ride-sharing. However, as we are considering urban mobility, we can assume that there are shared bikes provided from different independent providers, and therefore this mode of transport is always available for the users and the company does not have to plan the rebalancing on their own. 
Furthermore, the current model is restricted to a predetermined fixed sequence of tasks. This was considered as given from our practical partners. However, this restricts the model and prevents possible further savings that might benefit from this flexibility. Moreover, the model might profit from allowing changes of drivers on a trip. We note that this restriction was introduced based on information from our industry partners. They reported that there was very limited acceptance for handing over cars during a trip. However, in our future work we plan to address this aspect as well as more flexibility in the timing of user tasks. Lastly, another interesting aspect could be the consideration of uncertain and varying travel times. While time-dependent travel times would {not} require major changes to our approach, since we already work on a time expanded network, the consideration of uncertainty would require major changes in the design of the solution approach. Furthermore, an unexpected delay of the driver at any point would affect any later co-rider on the route, requiring changes in the schedule during the day, which should also be addressed in future work.

\section*{Acknowledgements}

This work was supported by the Climate and Energy Funds (KliEn) [grant number 853767]; and the Austrian Science Fund (FWF): P 31366.

\bibliographystyle{apacite}
\bibliography{main}

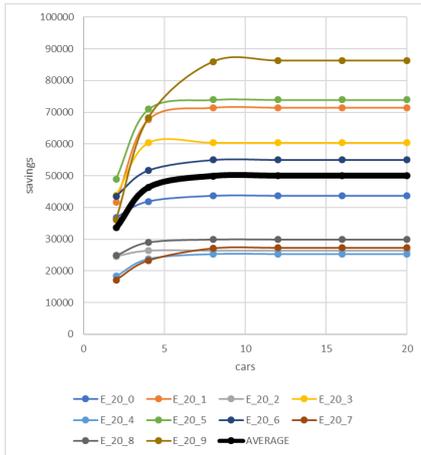
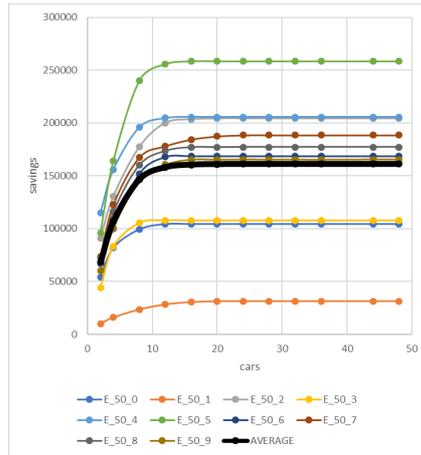
\end{document}